\documentclass[]{interact}

\usepackage{epstopdf}
\usepackage[caption=false]{subfig}

\usepackage{url}
\usepackage{color}

\usepackage[numbers,sort&compress]{natbib}
\bibpunct[, ]{[}{]}{,}{n}{,}{,}
\renewcommand\bibfont{\fontsize{10}{12}\selectfont}
\makeatletter
\def\NAT@def@citea{\def\@citea{\NAT@separator}}
\makeatother

\theoremstyle{plain}

\theoremstyle{definition}

\theoremstyle{remark}

\graphicspath{{./eps/}}

\begin{document}

\articletype{FULL PAPERS}

\title{Stochastic Approarch for Modeling the Soft Finger with the Creep Behavior}

\author{
\name{SUMITAKA Honji\textsuperscript{a}\thanks{CONTACT Sumitaka Honji. Email: honji@hcr.mech.kyushu-u.ac.jp},
  HIKARU Arita\textsuperscript{a} and KENJI Tahara\textsuperscript{a}}
\affil{\textsuperscript{a}Department of Mechanical Engineering, Kyushu University, 744 Moto'oka, Nishi-ku, Fukuoka, Japan}
}

\maketitle

\begin{abstract}
Soft robots have high adaptability and safeness which are derived from their softness, and therefore it is paid attention to use them in human society.
However, the controllability of soft robots is not enough to perform dexterous behaviors when considering soft robots as alternative laborers for humans.
The model-based control is effective to achieve dexterous behaviors.
When considering building a model which is suitable for control, there are problems based on their special properties such as the creep behavior or the variability of motion.
In this paper, the lumped parameterized model with viscoelastic joints for a soft finger is established for the creep behavior.
Parameters are expressed as distributions, which makes it possible to take into account the variability of motion.
Furthermore, stochastic analyses are performed based on the parameters' distribution.
They show high adaptivity compared with experimental results and also enable the investigation of the effects of parameters for robots' variability.
\end{abstract}


\section{Introduction}

\subsection{Problem setting}\label{sec:problem}

Recently, it's been paid attention to the use of robots in human living \cite{Lee2017}.
They are expected not only as labor but also as an augmented functionality like the Sixth-Finger \cite{Prattichizzo2014} or as a prosthetic leg, or hand.
In such an environment, robots are required to be safe for people or things around them.
Soft robots are composed of flexible materials such as polymer or fluid, therefore they show high adaptability to the environment.
For example, a soft gripper can grasp a flexible or unknown-shaped object stably by deforming along to its contour and then dispersing the grasping force.
In addition, soft robots can deform their body when an external force acts.
This means that they have good safety even if contact occurs suddenly.
Furthermore, soft robots are used in the field of biomimetics, which shows their high compatibility with living things.
From these points, soft robots have physical aptitudes in terms of human-robot interaction.

Although soft robots are required different functionalities from industrial robots in the context of human-robot interaction, they have not been realized yet.
Human space is not predefined unlike factories, so robots are required to act circumstantially.
For example, considering the object grasping task in our environment, robots may need to grasp and handle the object which can be placed in various attitudes.
Also, robots should control their power appropriately when dealing with fragile objects.
In such situations, robots should perform dexterously and realize highly efficient tasks by imitating human strategies.
However, the controllability of soft robots is not sufficient for realizing them.
At present, many soft robots are controlled by a simple controller such as Binary Control.
For example, a general pneumatic robot uses the solenoid binary valve to control the amount of inner fluid \cite{Polygerinos2017}.
Though this controller matches the adaptivity of soft robots, it is not suitable for performing dexterous motions since it may cause an oscillation of robots.
When considering realizing dexterous behaviors, a mechanical model-based controller would offer good performance \cite{Mengaldo2022}.
A model is useful not only for the feedforward control of robots but also for the feedback control by using a model as an observer.
Furthermore, by building a model, some mathematical tools such as stability analysis are available to evaluate a controller.
Many kinds of models have been studied and we mention some of them in the next subsection, but some challenges have not been discussed enough before.
First, non-linear properties emerge in the motion of soft robots.
The creep phenomenon is one of the representative characteristics of polymer material, in which a displacement does not converge but continues linearly when a constant external force is exerted.
Second, the motion is not uniquely determined for the same control input, but different per trial.
There are many factors such as fatigue, aging, temperature, humidity, the change of inner compositions, and so on, and they can affect the robot's properties such as viscoelasticity in a meaningful way.
Therefore, the modeling of soft robots is a challenging task.

To achieve dexterous behaviors by soft robots, the model-based control is effective and the model which is appropriate for real-time computation is necessary so that we can utilize it as an observer for example.
In this study, we work on the modeling of soft robots which contains (1) the creep behavior and (2) the variability of soft robots' motion.
The creep behavior is a feature that is present in soft robots' materials, so these motions can occur in many kinds of soft robots.
The variability also can have effects on the control.
Their mathematical expression of them makes it possible to evaluate the variable motion in qualitative and quantitative ways.
In addition, the model requires high solvability when used for real-time control.

\subsection{Related works}

Many types of models have been proposed not only for control but also for motion analysis of soft robots \cite{Mengaldo2022}.
Learning-based model is useful for dealing with the unknown behaviors of soft robots.
Giorelli et al. adopted the feedforward neural network (FNN) for learning the inverse kinematics of soft arm \cite{Giorelli2015}.
FNN showed a better result in the accuracy and computational cost of the Jacobian matrix compared with the numerical solution of the exact model.
Gupta et al. realized the dexterous manipulation which was learned from human demonstration using the approach based on reinforcement learning \cite{Gupta2016}.
Choi et al. introduced the 3D convolutional neural network to determine the grasping directions and the wrist orientations of the soft hand \cite{Choi2018}.
Though learning-based approaches are good for highly nonlinear systems, the stability analysis or the convergence proof is difficult because of its black box nature \cite{Thuruthel2018}.
Also, deep learning algorithms required a high-performance computer to deal with its significant computational loads, which could be a limitation for real-time application \cite{Kim2021}.
Though learning-based methods contain the uncertainty of soft robots in training data, the models which are obtained from learning algorithms are a black box and we cannot touch the physical mechanisms.
Mechanical models also have been studied from a lot of points of view.
The Finite Element Method (FEM) is one of the most representative tools for motion analysis of flexible objects because this method can be adaptable to complicated mechanisms such as soft robots with proper nodes and connections.
Kimura et al. proposed the constitutive law to express the rheological deformation and constructed the FEM simulation using it \cite{Kimura2003}.
To use FEM for control, the reduced order model is considered.
Duriez implemented the real-time simulation of the FEM model using SOFA, which is the open source simulation framework \cite{Allard2007}, then controlled the elastic soft robot in open-loop \cite{Duriez2013}.
Thieffry et al. proposed the control method based on the model reduction \cite{Thieffry2019},
and Katzschmann et al. performed the closed-loop control of the soft robotic arm by using FEM with model reduction \cite{Katzschmann2019}.
They also mentioned the computational limitation which came from the computing hardware.
Though FEM is a powerful tool, the rheological expression increases the computational cost and makes it hard to reduce or linearize the model.
Therefore, it is not suitable for real-time applications such as feedback control.
The Cosserat theory is one of the geometrically exact approaches based on continuous mechanics.
This theory is mainly used for the elongated structure such as the backbone of continuous manipulators or cables for tendon-driven robots.
Trivedi et al. used Cosserat rod theory to model the OctArm and showed that the accuracy of the model was better than that of the constant curvature model \cite{Trivedi2008}.
Renda et al. proposed the dynamical Cosserat model of the cable-driven octopus-like soft arm \cite{Renda2014}.
The comparison between the simulation and the experiment was confirmed, and the model showed an acceptable error concerning the length of the arm.
Adagolodjo et al. coupled the FEM of the soft robot and the Cosserat model of the driving cable, which makes the numerical simulation more stable \cite{Adagolodjo2021}.
Although the Cosserat approach performs well for robots with specific shapes, it is not suitable for full 3D deformations of soft robots because it drops some dimensions by special shape approximations \cite{Mengaldo2022}.
Therefore, this method can be a barrier against the dexterous motions of soft robots.
The lumped parameterized models are also proposed.
Su proposed Pseudorigid-Body 3R model to express the large deflection of cantilever beams \cite{Su2009},
and Huang et al. used this model for modeling the backbone of the continuum robot \cite{Huang2019}.
These models could reduce the computational cost.
Wang and Hirai approximated the deformation of the pneumatic fluidic elastomer actuator by the line-segment model \cite{Wang2017}.
Model parameters are identified using an optimization-based approach and the tip estimation is conducted.
Falkenhahn et al. presented the kinematics and dynamics of the continuum manipulator based on the Constant Curvature (CC) approach \cite{Falkenhahn2014}.
They showed that the CC model could be useful for trajectory planning and closed-loop controller design.
Santina et al. controlled the planar soft manipulator using the piecewise constant curvature model and realized the dynamical interaction with an environment \cite{Santina2018}.
However, these models do not match our soft fingers due to their creep behavior.
Therefore, the authors proposed the soft finger model which focused on the joint's creep displacement \cite{Honji2020}.
Though the Maxwell-type model showed a good result, this model is not enough to express the transient response.
In summary, the FEM or the Cosserat approach can represent the soft robots' behaviors more accurately than the lumped parameterized models, but utilizing them for real-time control is difficult because of their computational costs.
The lumped parameterized model is good in terms of this point.
However, the physical models mentioned above are based on the deterministic notion and they do not deal with the uncertainty of soft robots explicitly.

These modeling approaches above are based on the deterministic notion.
However, this methodology is not suitable for the variability of soft robots when controlling them dexterously because a deterministic model cannot express the various motion.
The variability of soft robots is bigger than rigid robots and it can cause troubles when controlling.
On the other hand, in the stochastic approach, the variability of the model which is contained in the input or model parameters, or other noise can be considered.
Generally, adding the noise term, which is from the Wiener process, the behavior of the system is analyzed by stochastic differential equations.
Although this approach is appropriate for considering the whole effects of uncertainties such as modeling errors, disturbances, sensor noises, and so on, each effect cannot be measurable.
Each effect is an important factor when analyzing or improving the behaviors of a system.
In this point, Cort\'{e}s et al. treated the first-order linear system stochastically using the random variable transformation \cite{Cortes2021}.
Model parameters are assumed to have stochastic distributions, then the stochastic analyses are conducted.
Our work is greatly inspired by their study.
To the best of our knowledge, a soft finger model that pays attention to its variability has not been proposed.
By extending the soft robot model from the stochastic term, the uncertainty of soft robots can be addressed mathematically.

\subsection{Proposal}

In this study, we propose a mathematical model which can express the creep behavior of the soft finger that we mentioned in Section~\ref{sec:problem}.
For the creep behavior, we combine viscous and elastic elements.
It is shown that the linear combination of spring and damper is efficient for this modeling.
Then with the lumped parameterized model, the whole dynamics is derived.
Concerning the variability of soft robots' motion, the stochastic approach is introduced.
By assuming model parameters as stochastic distributions, the solution of the model can be represented as the stochastic process.
In particular, the variability of the finger deformation is explained by the stochastic parameters.
Furthermore, the sensitivity analysis is performed to show which parameter has more influence on the variability of a soft finger.

The contribution of this study is as follows.
The proposed model is constructed with the lumped parameterized model and the linear 3-element viscoelastic joint model and has the advantage of the computational cost, so it is expected to utilize for the model-based control.
This model can be adaptable to other types of soft robots such as continuum robots or compliant mechanisms with the appropriate link system expression and the viscoelastic component.
Furthermore, the stochastic model makes it possible to evaluate the uncertainty such as the variability of motion which comes from the change of viscoelastic parameters qualitatively and quantitatively. 
For example, how much the material affects is an important factor in the design process.

The construction of this paper is as follows.
In Section~\ref{sec:modeling}, we derive the mathematical model of the soft finger using the lumped parameterized approximation.
The analytical solution of the joint deformation is formulated in Section~\ref{sec:solution}, which is used for the parameter estimation and stochastic analysis later.
The parameter estimation is conducted in Section~\ref{sec:parameter}.
Parameters are obtained as stochastic distributions, this is the key point of this study.
In Section~\ref{sec:analysis}, the efficiency of the proposed modeling is shown by the stochastic analysis.
Through the discussion in Section~\ref{sec:discussion}, we conclude our proposal.

\section{Lumped parameterized modeling of a soft finger}\label{sec:modeling}

\subsection{Design of the soft finger}

\begin{figure}[tb]
\centering
\includegraphics[width=\linewidth]{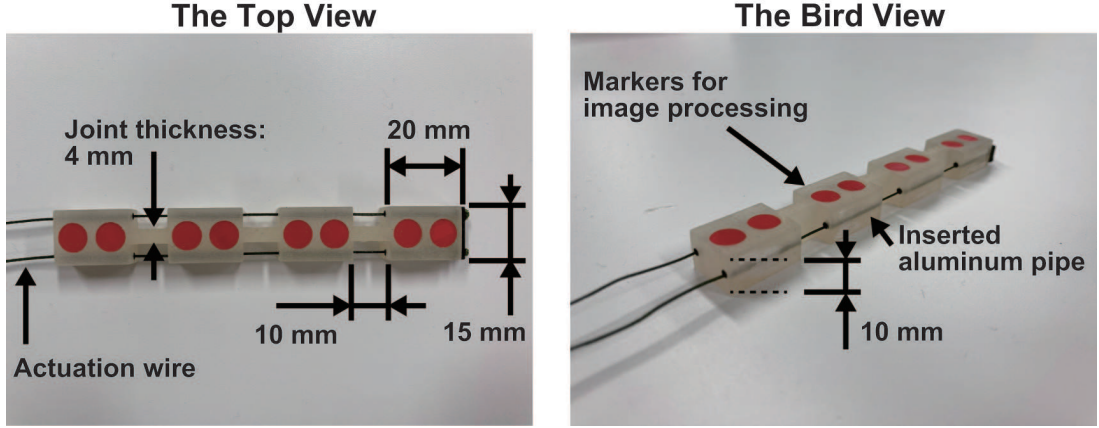}
\caption{The geometry of the soft finger.}
\label{fig:finger}
\end{figure}

In this section, the mathematical model of soft robots is derived.
The lumped parameterized model is selected due to its computational benefit, which will be useful for real-time control in the future.
Also, we choose the soft finger shown in Figure~\ref{fig:finger} as a modeling target.
The shape is suitable for this modeling because the finger is constricted and can easily be fitted with links and joints.
Of course, the lumped parameterized model is useful for other types of soft robots, for example, adding links and joints is appropriate for continuum manipulators.
The characteristics of this finger are as follows.
It is easy to build it by 3D printing because it is integrally molded.
Another advantage of this design is that the change in the joint thickness is directly related to the change in finger stiffness.
This means that we can intuitively realize appropriate fingers for various purposes.

This finger is formed by the 3D printer (AGILISTA3200, KEYENCE CORP. \cite{keyence}) whose material is high-hardness silicone rubber, which is a product also by KEYENCE (AR-G1H).
The finger is actuated by two antagonistic wires, one of which passes through the flexion side, and the other goes through the extension side, and both wires are fixed at the tip of the finger.
In this case, the degree of freedom of joint space is 3 and that of actuator space is 2, which means actuation wires cannot control the internal force or the stiffness of the finger.
PE fishing line made of Ultra Dyneema fiber is chosen as the actuation wire.
To reduce the friction between the wires and the finger, and to prevent the finger from being torn by the wire shear stress, aluminum tubes are inserted into the wire paths.
This also makes the linked approach more reasonable.
These two wires are driven by DC motors located at the base of the finger.

\subsection{Dynamic model}
\begin{figure}
\centering
\subfloat[Lumped parameterized model.]{%
\resizebox*{7cm}{!}{\includegraphics{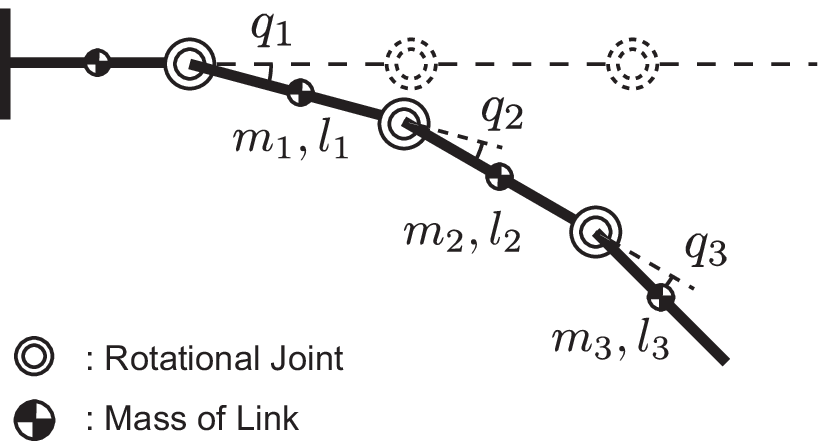}}}\hspace{5pt}
\subfloat[Joint structure.]{%
\resizebox*{7cm}{!}{\includegraphics{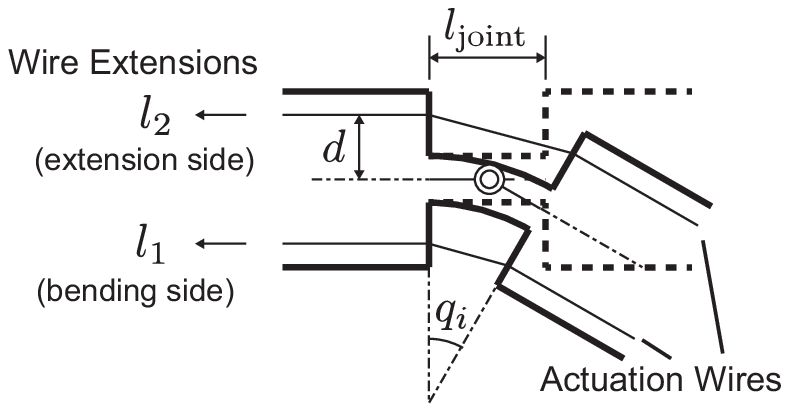}}}
\caption{The model of a soft finger and the detail structure of joint.} \label{fig:model}
\end{figure}

The dynamic model of our soft finger is described by the lumped parameterized model.
The modeling is performed in 2 phases.
In this section, the serial link model is derived without viscoelastic relationships.
The viscoelastic feature such as the creep behavior is considered in the next section as the joint model.
The geometric model of the one finger is shown in Figure~\ref{fig:model}(a).
To derive the mathematical model, we used the Lagrange method.
The finger is assumed to move in the horizontal plane so that the effect of gravity is ignored, so the Lagrangian $L$ is equal to the kinetic energy and is as follows:
\begin{equation}
L = \sum_{i = 1}^{3} \left( \frac{1}{2}m_i v_i^2 + \frac{1}{2}I_i \left( \sum_{j=1}^{i} \dot{q}_j \right)^2 \right). \label{eq:lagrangian}
\end{equation}
Here, $m_i$ is the mass of each member, $v_i \left( = \sqrt{\dot{x}_i^2+\dot{y}_i^2} \right)$ is the velocity of each mass point, $I_i$ is the moment of inertia and $q_i$ expresses the relative angle from the previous member.
Then the equation of motion is expressed as,
\begin{equation}
\bm M \left( \bm q \right) \ddot{\bm q} +  \bm h \left( \bm q,\dot{\bm q} \right) + \bm \tau_\mathrm{in} = \bm \tau_\mathrm{ac}, \label{eq:fingerdyn}
\end{equation}
where $\bm q = \left[ q_1 \quad q_2 \quad q_3 \right]^\top$ is the generalized vector, $\bm M\left(\bm q\right)$ is the inertia matrix, $\bm h\left(\bm q,\dot{\bm q}\right)$ is a vector of the nonlinear term and $\bm \tau_\mathrm{ac}$ is the actuating torque acting in the joint space.
Also, $\bm \tau_\mathrm{in}$ is the vector of joint internal torque generated by the joint viscoelastic element, which will be detailed in the next chapter.
Actuating torque caused by wires can be obtained from the principle of virtual work \cite{Murray2017}:
\begin{equation}
\bm \tau_\mathrm{ac} = \bm P \bm f = \left( \frac{\partial \bm l}{\partial \bm q} \right)^\top \bm f. \label{eq:actuationtorque}
\end{equation}
Here, $\bm P$ is the wire Jacobian matrix, $\bm l = \left[ l_1 \quad l_2 \right]^\top$ is the wire elongation vector and ${\bm f} \in \mathbb{R}^2$ is the wire tension vector.
For our finger design, the wire extensions of one-link fingers are calculated from the geometric relationship shown in Figure~\ref{fig:model}(b).
In this figure, $d$ represents the distance between the major axis of the linked part and the wire path, and $l_\mathrm{joint}$ is the length of the joint part without bending.
The wire extensions of the joint $i$ are shown below:
\begin{align}
l_1 &= l_\mathrm{joint} - 2 \sqrt{\left(\frac{l_\mathrm{joint}}{2} \right)^2 + d^2} \cos{ \left( \frac{q_i}{2} + a \right) }, \label{eq:bendjoint} \\
l_2 &= l_\mathrm{joint} - 2 \sqrt{\left(\frac{l_\mathrm{joint}}{2} \right)^2 + d^2} \cos{ \left( \frac{q_i}{2} - a \right) }, \label{eq:extendjoint} \\
a &= \tan^{-1} { \frac{2d}{l_\mathrm{joint}} }. \nonumber
\end{align}
Since our finger design is the three-link finger, each wire length is the sum of eq.(\ref{eq:bendjoint}) or eq.(\ref{eq:extendjoint}) and represented as,
\begin{align}
l_1 &= 3 l_\mathrm{joint} - 2 \sqrt{\left(\frac{l_{\rm joint}}{2} \right)^2 + d^2} \sum_{i=1}^3 \cos{ \left( \frac{q_i}{2} + a \right) }, \label{eq:bendwire} \\
l_2 &= 3 l_\mathrm{joint} - 2 \sqrt{\left(\frac{l_{\rm joint}}{2} \right)^2 + d^2} \sum_{i=1}^3 \cos{ \left( \frac{q_i}{2} - a \right) }. \label{eq:extendwire}
\end{align}
%

%

\subsection{Viscoelastic constitutive equation}

\begin{figure}[tb]
\centering
\includegraphics[width=0.5\linewidth]{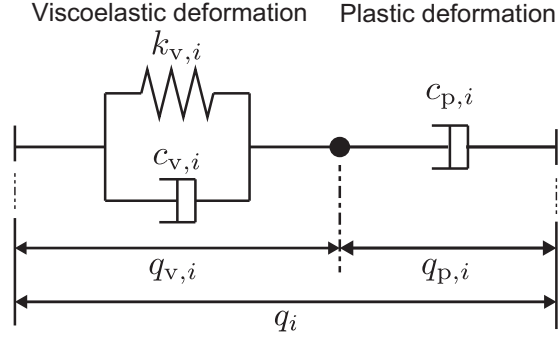}
\caption{Viscoelastic 3 elements model of joint $i$.}
\label{fig:joint}
\end{figure}

In our previous work, the joint was expressed as the Maxwell model \cite{Honji2020}.
Though this model could explain the creep phenomenon, it was not sufficient for the rheological deformation.
The minimal elements for this motion are constituted by the combination of 1 spring and 2 dampers \cite{Kimura2003}.
Therefore, the 3-element model which contains 1 linear spring and 2 linear dampers is adopted for each joint, as shown in Figure~\ref{fig:joint}.
A parallel set of a spring and a damper and another damper are combined, the former expresses the viscoelastic behavior and the latter does the plastic deformation.
Plastic deformation is one of the important features of polymer materials and existed in our finger.

Here, the viscoelastic relationship is derived.
For each joint, 3 equations are satisfied:
\begin{eqnarray}
\tau_i &=& k_{\mathrm{v},i} q_{\mathrm{v},i} + c_{\mathrm{v},i} \dot{q}_{\mathrm{v},i}, \label{eq:viscoelastic1} \\
\tau_i &=& c_{\mathrm{p},i} \dot{q}_{\mathrm{p},i}, \label{eq:viscoelastic2} \\
q_i &=& q_{\mathrm{v},i} + q_{\mathrm{p},i}, \label{eq:viscoelastic3}
\end{eqnarray}
where $k_{\mathrm{v},i}, c_{\mathrm{v},i}, c_{\mathrm{p},i}$ are viscoelastic coefficients, $q_{v,i}$ is the displacement of viscoelastic part and $q_{p,i}$ is that of plastic part of joint $i$.
From eq.(\ref{eq:viscoelastic1}) $\sim$ (\ref{eq:viscoelastic3}), the viscoelastic relation is derived as,
\begin{equation}
c_{\mathrm{v},i} c_{\mathrm{p},i} \dot{q}_i + c_{\mathrm{p},i} k_{\mathrm{v},i} q_i = \left( c_{\mathrm{v},i} + c_{\mathrm{p},i} \right) \tau_i + k_{\mathrm{v},i} \int \tau_i dt .\label{eq:viscoelastic}
\end{equation}

Extending eq.(\ref{eq:viscoelastic}) to all joints gives the constitutive equation as follows:
\begin{eqnarray}
\bm A \dot{\bm q} + \bm B \bm q &=& \bm C \bm \tau + \bm D \int \bm \tau dt, \label{eq:jointdyn} \\
\bm A &=& \mathrm{diag} \left[ c_{\mathrm{v},i} c_{\mathrm{p},i} \right], \nonumber \\
\bm B &=& \mathrm{diag} \left[ c_{\mathrm{p},i} k_{\mathrm{v},i} \right], \nonumber \\
\bm C &=& \mathrm{diag} \left[ c_{\mathrm{v},i} + c_{\mathrm{p},i} \right], \nonumber \\
\bm D &=& \mathrm{diag} \left[ k_{\mathrm{v},i} \right]. \nonumber
\end{eqnarray}

Finally, the whole body dynamics is constructed from eq.(\ref{eq:fingerdyn}) and eq.(\ref{eq:jointdyn}) as,
\begin{equation}
\begin{cases}
\bm M \left( \bm q \right) \ddot{\bm q} +  \bm h \left( \bm q,\dot{\bm q} \right) + \bm \tau_{\rm in} &= \bm \tau_{\rm ac} \\
\bm A \dot{\bm q} + \bm B \bm q &= \bm C \bm \tau_{\rm in} + \bm D \int \bm \tau_{\rm in} dt
\end{cases}. \label{eq:dynamics}
\end{equation}
%

\section{Analytical solution of the joint displacement}\label{sec:solution}

\subsection{Joint behavior for step input}

The analytical solution of the joint displacement plays an important role not only in the parameter estimation but also in the stochastic analysis as shown in Section~\ref{sec:parameter} and Section~\ref{sec:analysis} respectively.
However, it's hard to describe the analytical deformation for general inputs.
That's why, some assumptions are imposed.
First, the effects of the inertia and the nonlinear term can be ignored because of the lightness of the finger.
This means that the joint level eq.(\ref{eq:viscoelastic}) dominates the joint displacement.
Also, the joint torque is referred to as constant when the actuation wire is pulled by a constant force.
The adequacy of these assumptions is confirmed through the comparison with simulational and experimental results in the next section.
Under these assumptions, the analytical solution is derived.
For simplicity, we drop the subscript $i$ which implies the joint number because every joint has the same formulation.
With $\tau$ as input and $q$ as output, the transfer function $G$ of eq.(\ref{eq:viscoelastic}) is obtained as,
\begin{equation}
G \left( s \right) = \frac{\mathcal{L}\left[ q \right]}{\mathcal{L}\left[ \tau \right]}
= \frac{\left(c_\mathrm{v} + c_\mathrm{p} \right) s + k_\mathrm{v}}{c_\mathrm{v} c_\mathrm{p} s^2 + c_2 k_\mathrm{v} s} .
\end{equation}

For the step input, the joint behavior is explicitly expressed by viscoelastic parameters.
In this case, considering a step torque of magnitude $K$,
\begin{align}
\mathcal{L}\left[ q \right] &= G \cdot \frac{K}{s} , \\
q &= \frac{K}{k_\mathrm{v}} \left\{ 1 - \exp{\left( -\frac{k_\mathrm{v}}{c_\mathrm{v}}t \right)} \right\}  + \frac{K}{c_\mathrm{p}}t + q_\mathrm{ini} ,\label{eq:jointmodel}
\end{align}
where $t$ is the time from when the input starts to work.
The first term on the right side expresses the elastic deformation caused by the parallel viscoelastic part and the second one means the plastic deformation.
Please note that the initial displacement is added as the offset.

\section{Stochastic expression of viscoelastic parameters}\label{sec:parameter}

\subsection{Experimental setup}

To analyze the viscoelastic model with the actual value, parameter estimation should be performed.
The viscoelastic parameters are obtained as distributions for the stochastic extension.

\begin{figure}[tb]
\centering
\includegraphics[width=\linewidth]{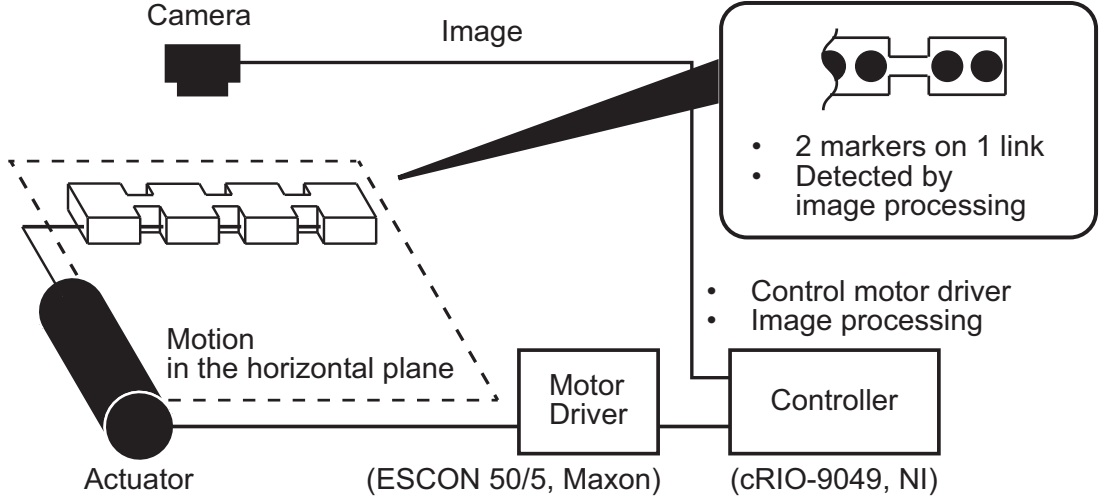}
\caption{Abstract of experimental setup.}
\label{fig:setup}
\end{figure}

Figure~\ref{fig:setup} shows the experimental setup for parameter estimation.
The soft finger is actuated by one wire and moves in the air only on the horizontal plane.
2 markers are attached to each link and move with the soft finger.
The motion of the finger is captured by a camera (Realsense D435, INTEL CORP. \cite{intel}), then a camera image is processed on the real-time controller (cRIO-9049, NATIONAL INSTRUMENTS CORP. \cite{ni}) to detect the positions of the markers and calculate a tilt of each link.

In this experiment, a constant wire force was applied to the finger for 30 seconds and the deformation was recorded.
This procedure was repeated 100 times.
Deterministic parameters were estimated for each procedure, which meant that we obtained 100 estimated values for each parameter.

\subsection{Parameter distribution}

Estimated viscoelastic parameters for each trial are different, but regular impacts due to its fatigue were not confirmed.
In general, a soft robot such as a soft finger shows different behavior in each trial for the same input.
In this study, the cause of this phenomenon is assumed that the parameters in the dynamics vary, which are not deterministic values.
As a model case, the soft finger is driven for a rather short time, and the distribution of the parameters is revealed.

\begin{figure}
\centering
\subfloat[Viscosity $c_\mathrm{v}$ of joint 1.]{%
\resizebox*{4.5cm}{!}{\includegraphics{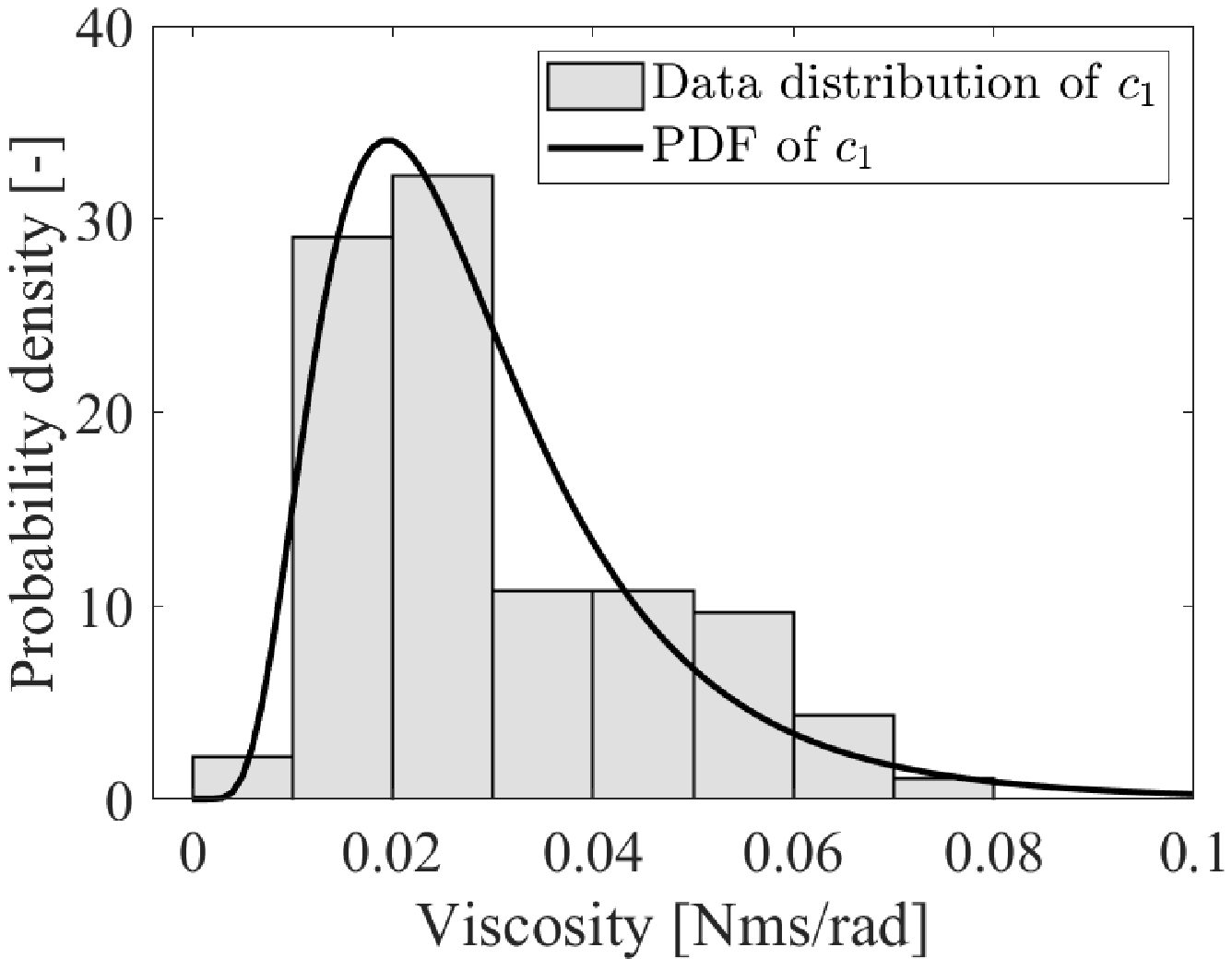}}}\hspace{5pt}
\subfloat[Viscosity $c_\mathrm{p}$ of joint 1.]{%
\resizebox*{4.5cm}{!}{\includegraphics{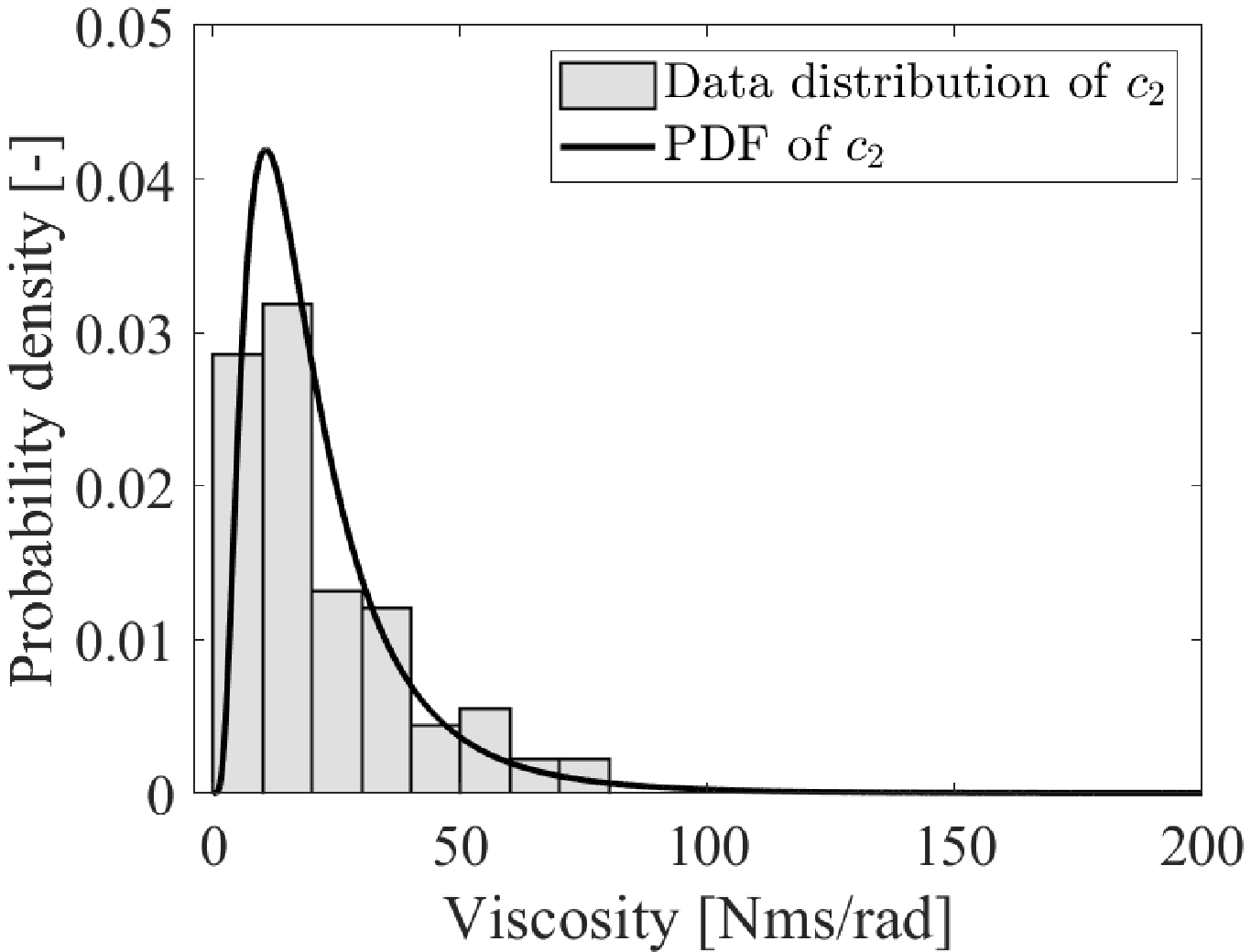}}}\hspace{5pt}
\subfloat[Elasticity $k_\mathrm{v}$ of joint 1.]{%
\resizebox*{4.5cm}{!}{\includegraphics{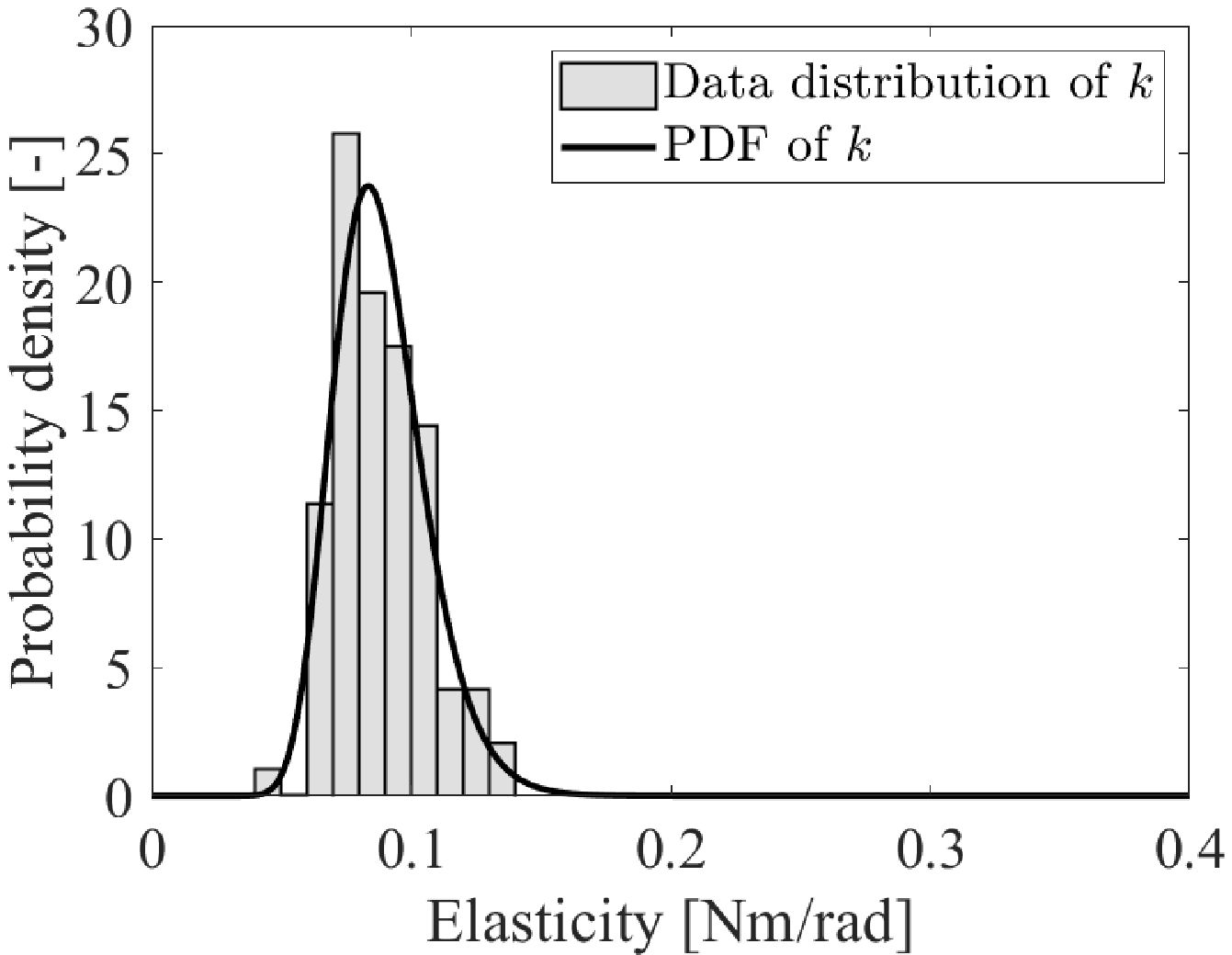}}}\hspace{5pt}
\subfloat[Viscosity $c_\mathrm{v}$ of joint 2.]{%
\resizebox*{4.5cm}{!}{\includegraphics{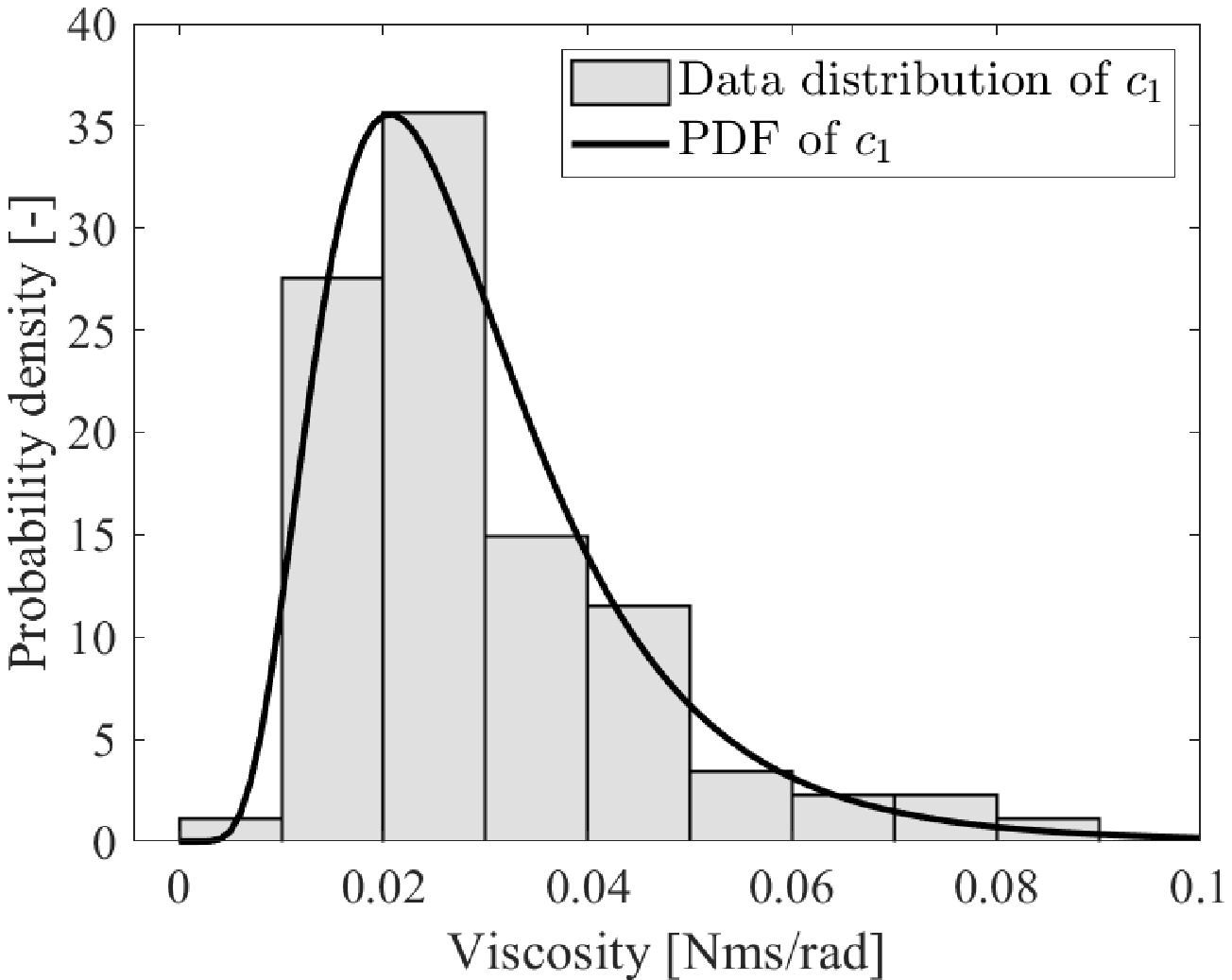}}}\hspace{5pt}
\subfloat[Viscosity $c_\mathrm{p}$ of joint 2.]{%
\resizebox*{4.5cm}{!}{\includegraphics{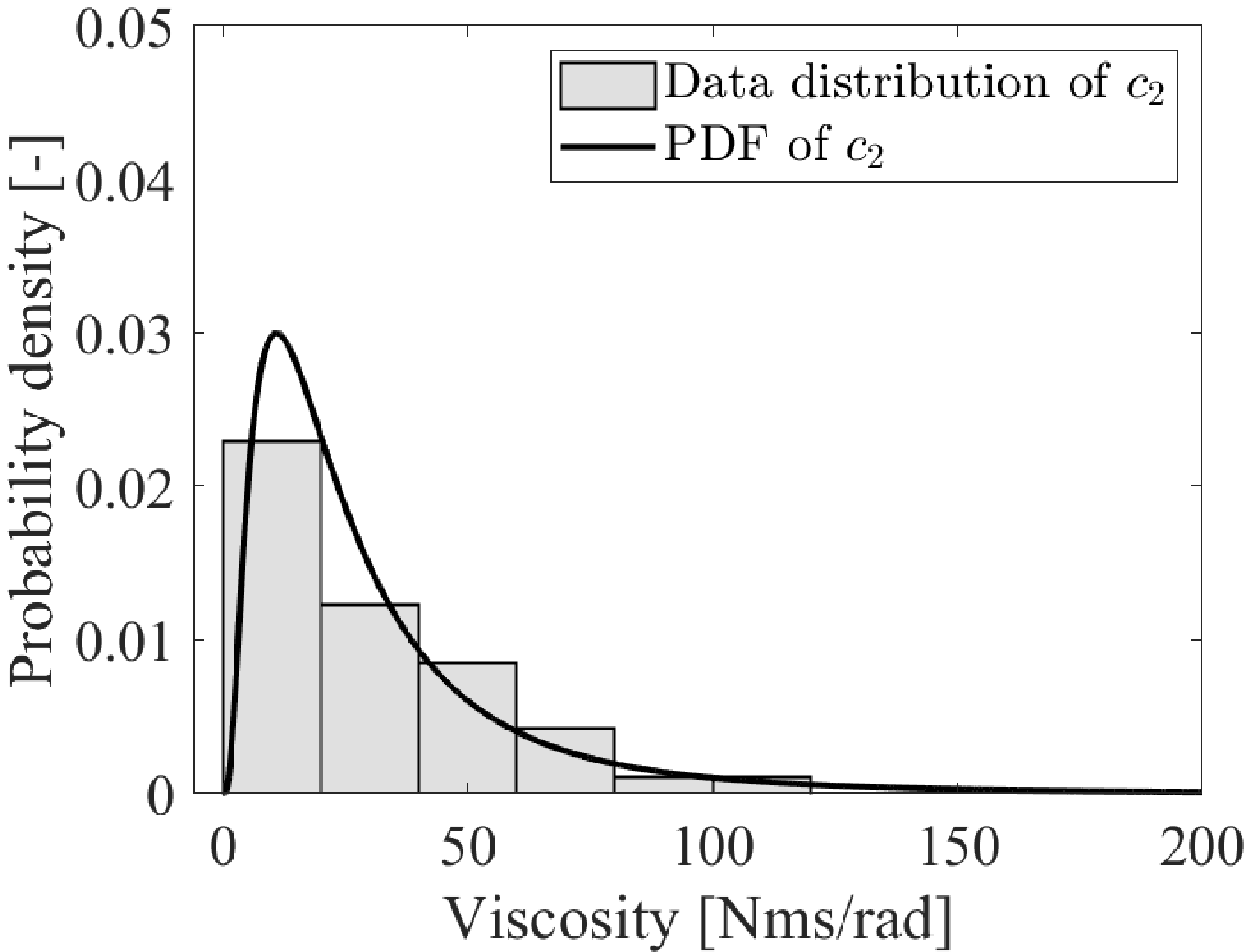}}}\hspace{5pt}
\subfloat[Elasticity $k_\mathrm{v}$ of joint 2.]{%
\resizebox*{4.5cm}{!}{\includegraphics{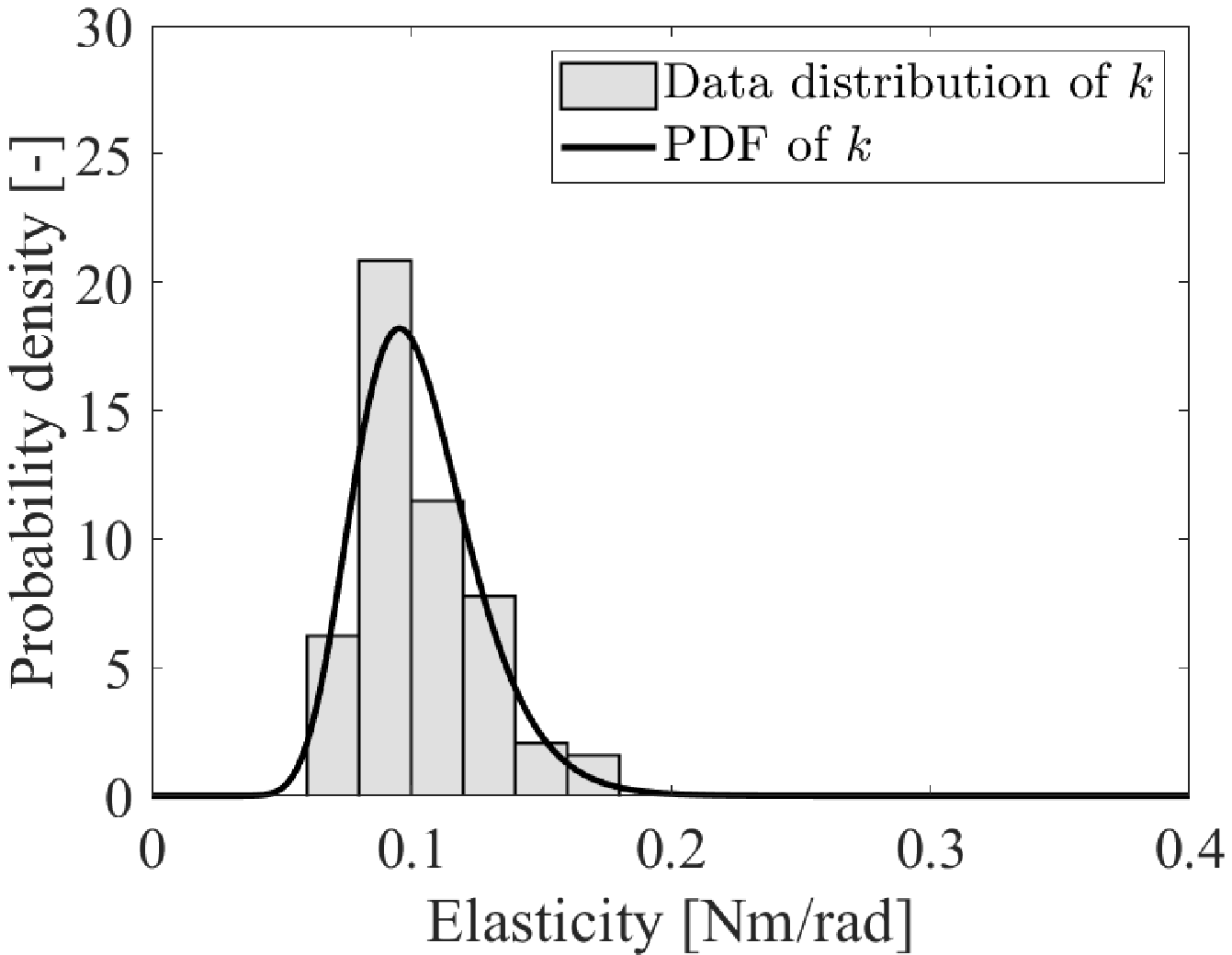}}}\hspace{5pt}
\subfloat[Viscosity $c_\mathrm{v}$ of joint 3.]{%
\resizebox*{4.5cm}{!}{\includegraphics{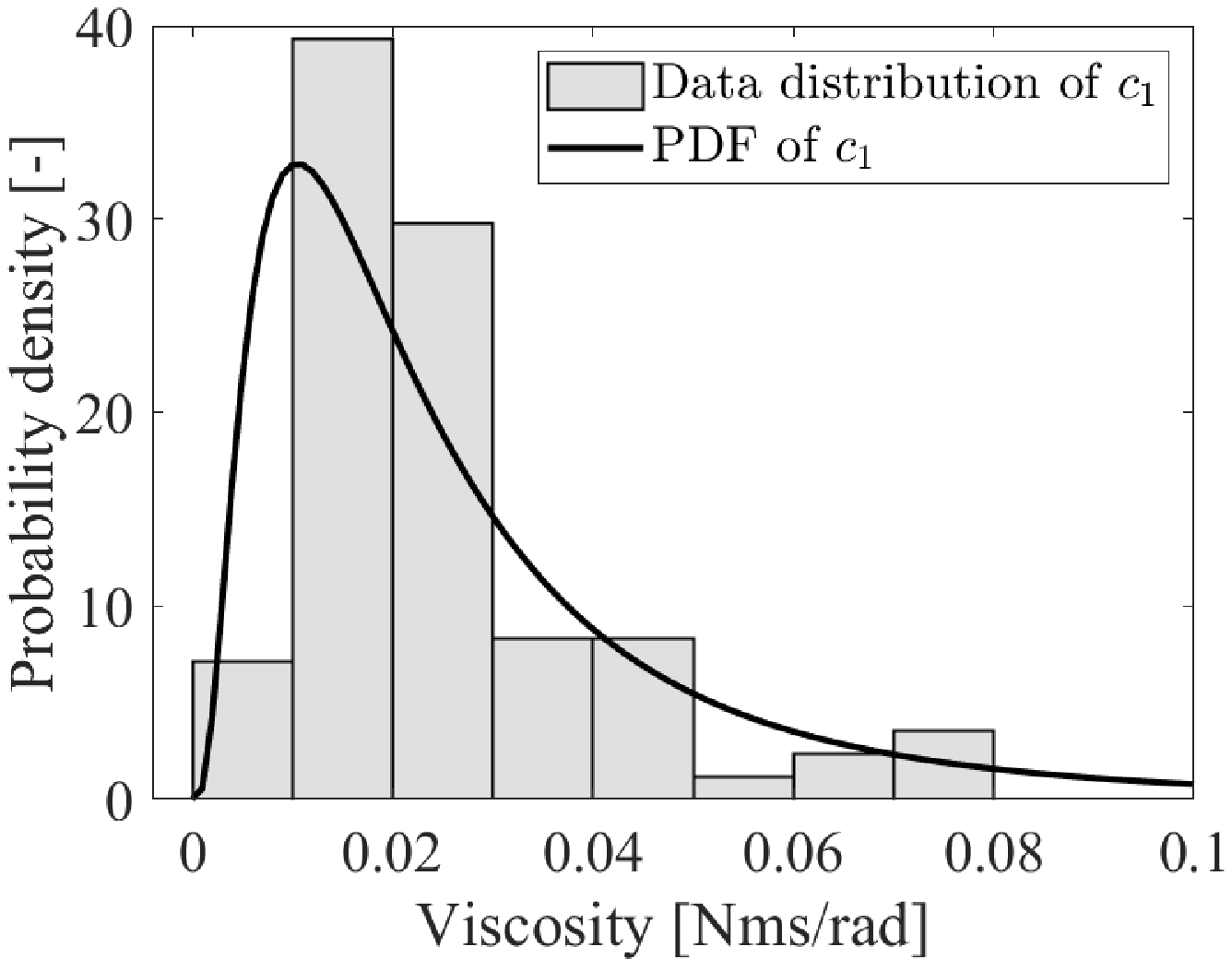}}}\hspace{5pt}
\subfloat[Viscosity $c_\mathrm{p}$ of joint 3.]{%
\resizebox*{4.5cm}{!}{\includegraphics{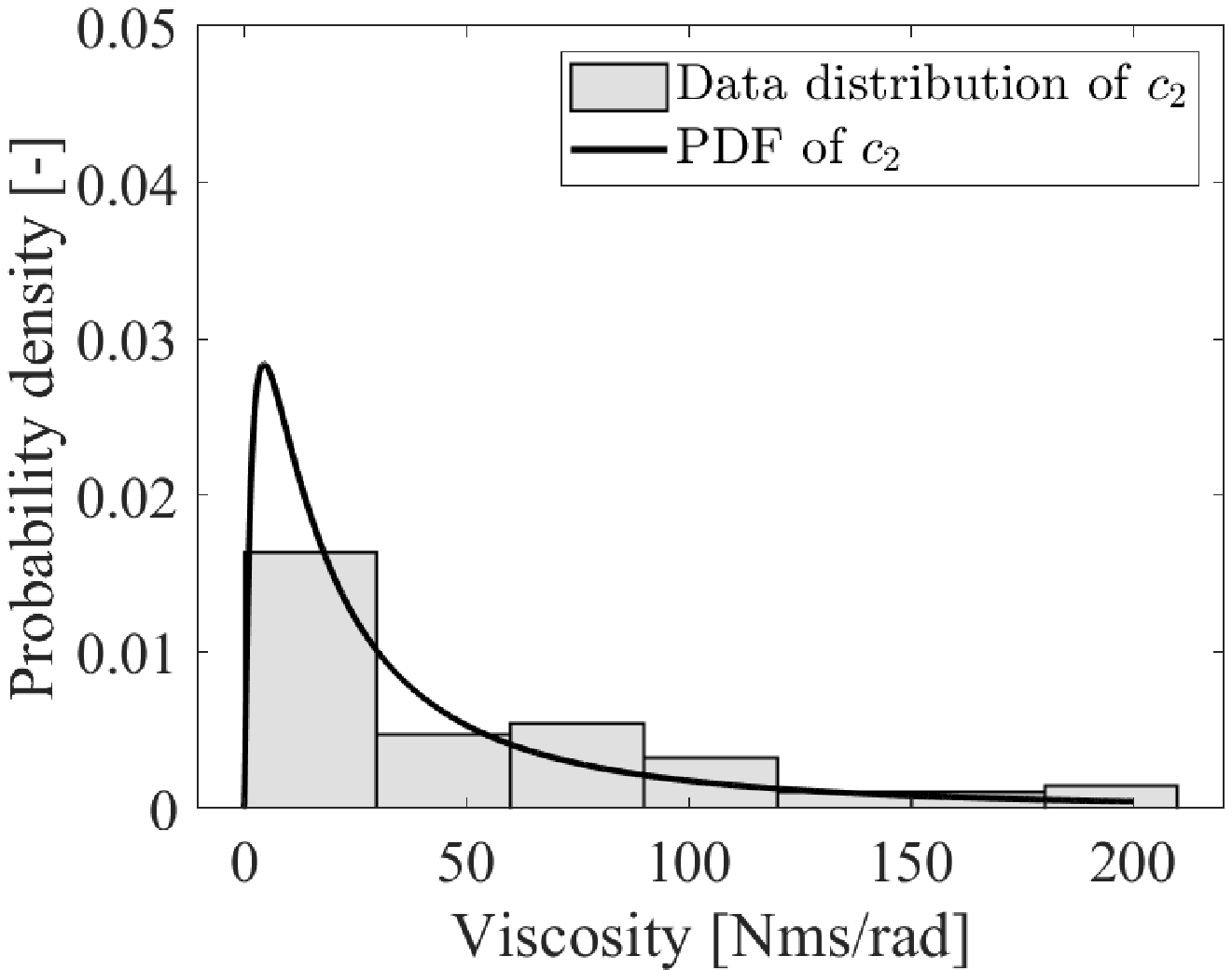}}}\hspace{5pt}
\subfloat[Elasticity $k_\mathrm{v}$ of joint 3.]{%
\resizebox*{4.5cm}{!}{\includegraphics{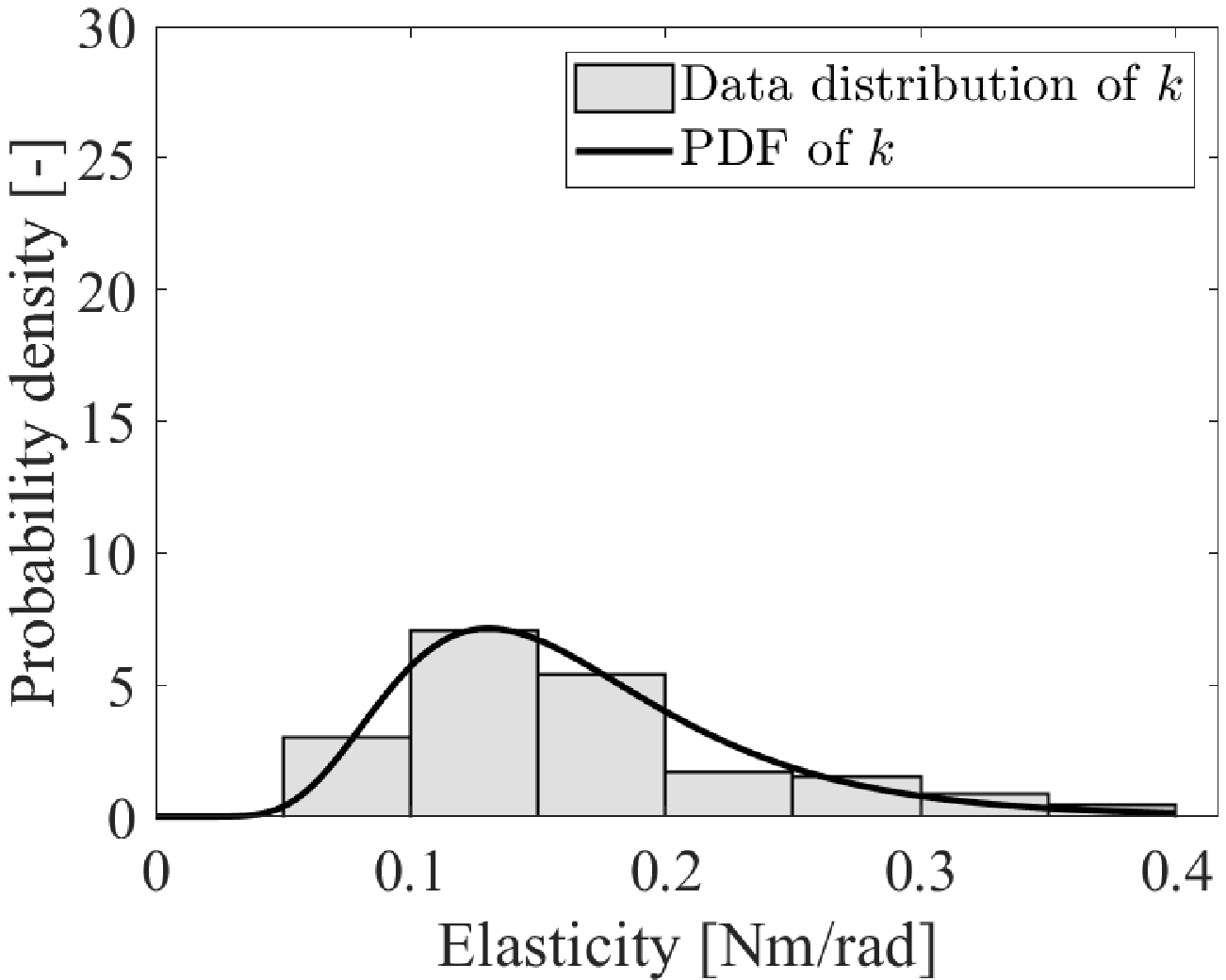}}}\hspace{5pt}
\caption{
The histograms and the PDFs are shown. The row shows the result of viscoelatic parameters of a joint, and the column does that of the same element of every joint.
} \label{fig:dist}
\end{figure}

From the experimental result, a left-balanced distribution for each viscoelastic parameter was confirmed as shown in Figure~\ref{fig:dist}.
Note that we outlier values that are 3 times away from the median absolute deviation or negative ones.
Also, these histograms are normalized so that the total area equals 1.
Due to this normalization, we can compare histograms with obtained Probability Density Functions (PDF).
From the shape of the histograms, the log-normal distribution is selected.
The probability function of the log-normal distribution for the parameter $x$ is defined as,
\begin{equation}
f \left( x,\sigma \right) = \frac{1}{x \sigma \sqrt{2 \pi}} \exp{ \left\{ \frac{\left( \log{x} - \mu \right)^2}{2 \sigma^2} \right\}}.
\end{equation}
The \texttt{fitdist} function (available in MATLAB) is used to find the shape parameters $\sigma$, $\mu$.
PDFs of each joint and parameter are also shown in Figure~\ref{fig:dist}, and the obtained shape parameters are shown in the Table.\ref{table:pdf param}.
The shape parameters are hyperparameters and there is no engineering meaning for the parameter itself.
From this result, the parallel viscosity $c_\mathrm{v}$ has similar shape distribution regardless of its joint position.
On the other hand, the series viscosity $c_\mathrm{p}$ is different.
The closer the joint is to the base, the higher and narrower the shape of the distribution becomes.
This is because the displacement of joint 3 is small, and also the measurement error from image processing accumulates.
The distributions of $k_\mathrm{v}$ are different at joints.
The shape of joint 3 is low and wide, which comes from the difficulty of the measurement.


\begin{table}[tbp]
  \centering
  \tbl{Shape parameters of PDF}
  {\begin{tabular}{cccccccccc}
  \toprule
    	&	& $c_\mathrm{v}$	&	&	& $c_\mathrm{p}$	&	&	& $k_\mathrm{v}$	& \\ \cmidrule(r){2-4} \cmidrule(r){5-7} \cmidrule(r){8-10}
    	& Joint 1	& Joint 2	& Joint 3	& Joint 1	& Joint 2	& Joint 3	& Joint 1	& Joint 2	& Joint 3 \\ \cmidrule(r){2-4} \cmidrule(r){5-7} \cmidrule(r){8-10}
    $\sigma$	& 0.5232	& 0.4838	& 0.8223	& 0.7011	& 0.8496	& 3.2316	& 0.1973	& 0.2238	& 0.3957 \\
    $\mu$	& -3.6635	& -3.6476	& -3.8777	& 2.8538	& 3.1129	& 1.3114	& -2.4441	& -2.2986	& -1.8800 \\ \bottomrule
  \end{tabular}}\label{table:pdf param}
\end{table}

\section{Stochastic analysis}\label{sec:analysis}

\subsection{Stochastic expression of joint angle transition}

\begin{figure}
\centering
\subfloat[Initial angle $q_\mathrm{ini}$ of joint 1.]{%
\resizebox*{4.5cm}{!}{\includegraphics{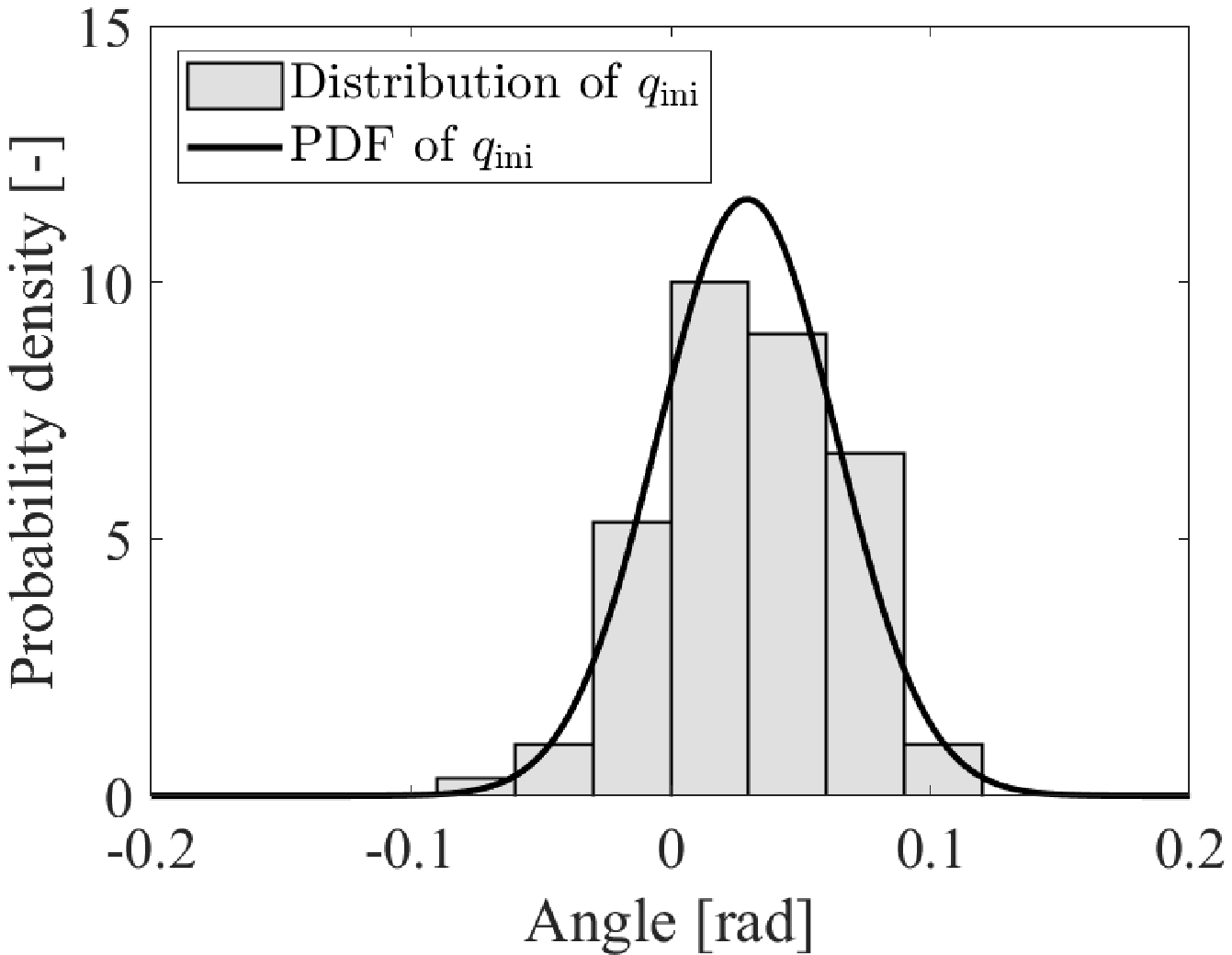}}}\hspace{5pt}
\subfloat[Initial angle $q_\mathrm{ini}$ of joint 2.]{%
\resizebox*{4.5cm}{!}{\includegraphics{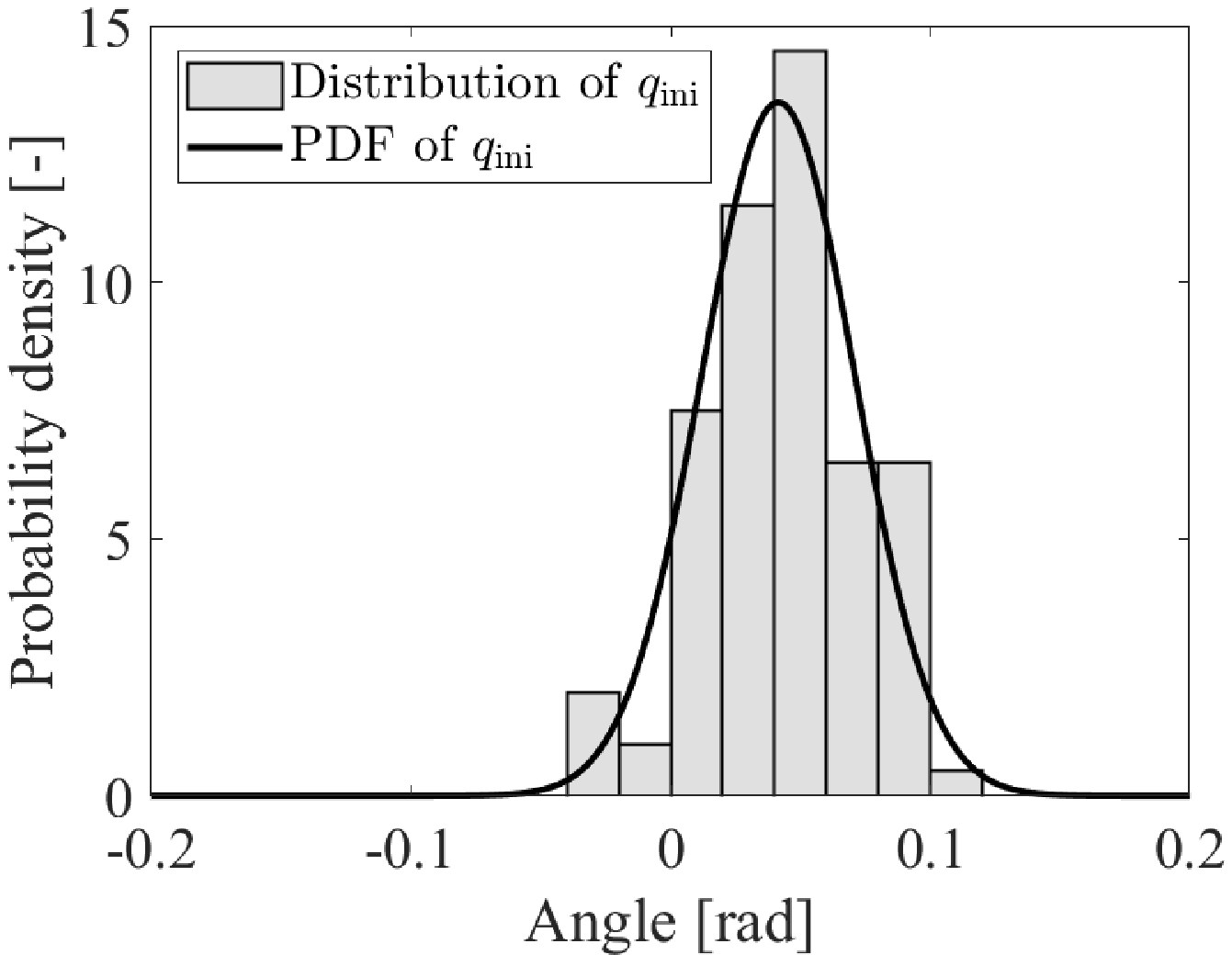}}}\hspace{5pt}
\subfloat[Initial angle $q_\mathrm{ini}$ of joint 3.]{%
\resizebox*{4.5cm}{!}{\includegraphics{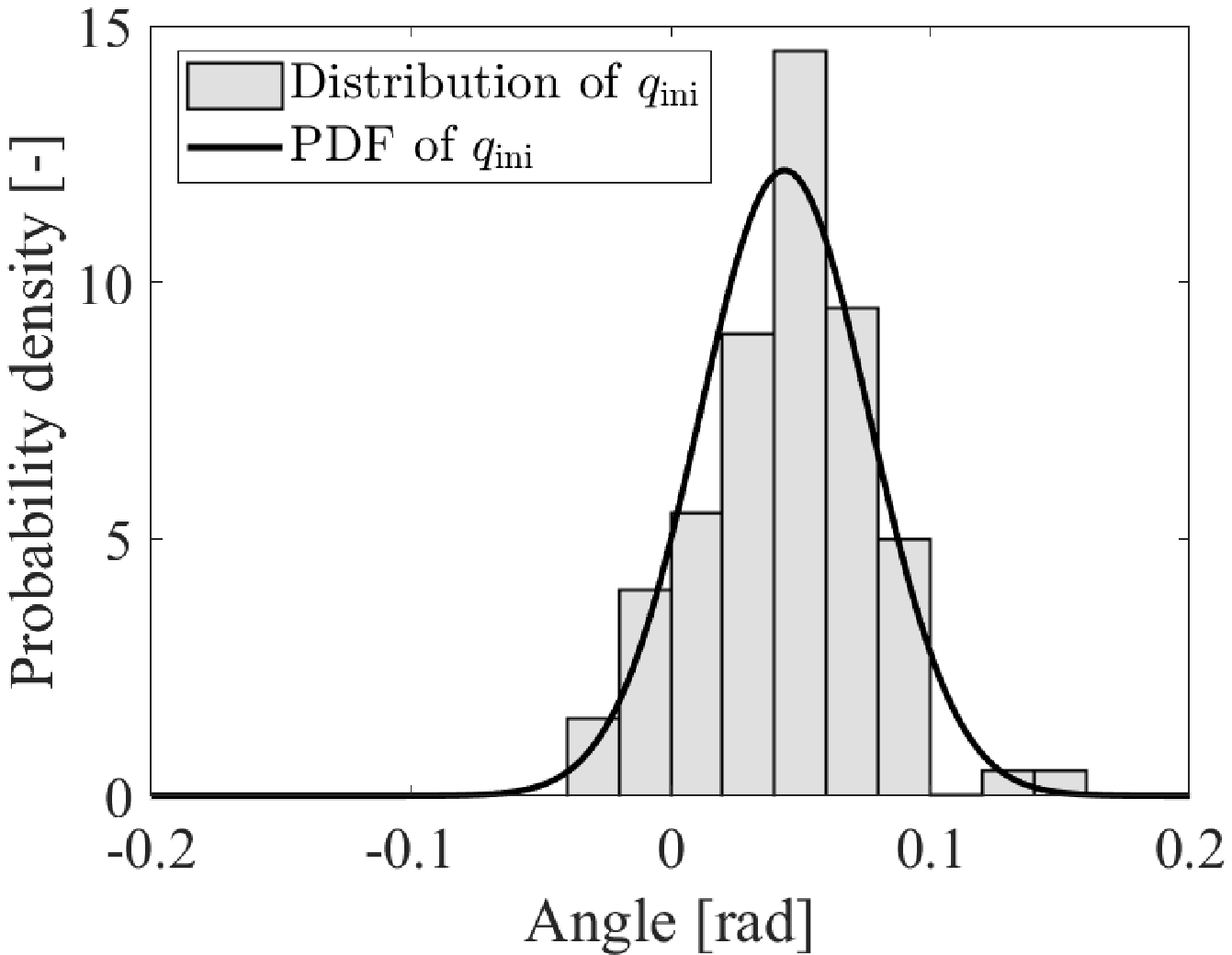}}}\hspace{5pt}
\caption{
The histograms and the PDFs of the initial angle are shown.
} \label{fig:dist_ini}
\end{figure}

By considering the distribution of viscoelastic parameters, we can evaluate how much the joint angle fluctuates from the stochastic analysis.
Here, the Random Variable Transformation method\cite{Soong1973} is used to obtain the PDF of each joint angle.
Assuming that each distribution is independent, the joint PDF $f_{c_\mathrm{v},c_\mathrm{p},k_\mathrm{v},q_\mathrm{ini}}$ is expressed as follows:
\begin{equation}
f_{c_\mathrm{v},c_\mathrm{p},k_\mathrm{v},q_\mathrm{ini}} \left( c_\mathrm{v}, c_\mathrm{p}, k_\mathrm{v}, q_\mathrm{ini} \right) = f_{c_\mathrm{v}} \left( c_\mathrm{v} \right) f_{c_\mathrm{p}} \left( c_\mathrm{p} \right) f_{k_\mathrm{v}} \left( k_\mathrm{v} \right) f_{q_\mathrm{ini}} \left( q_\mathrm{ini} \right),
\end{equation}
where $f_{c_\mathrm{v}}, f_{c_\mathrm{p}}, f_{k_\mathrm{v}}, f_{q_\mathrm{ini}}$ are the PDF of $c_\mathrm{v}$, $c_\mathrm{p}$, $k_\mathrm{v}$ and $q_\mathrm{ini}$, respectively.
The PDFs of the viscoelastic parameters $c_\mathrm{v}, c_\mathrm{p}, k_\mathrm{v}$ are obtained in the previous section, and the PDF of the initial joint angle $q_\mathrm{ini}$ is also derived from the experimental result as the normal distribution in Figure~\ref{fig:dist_ini}.
Now, we define 2 random vectors, $\bm{u} = \left[\begin{array}{cccc} c_\mathrm{v} & c_\mathrm{p} & k_\mathrm{v} & q_\mathrm{ini}\end{array} \right]^\top$ contains known random variables and $\bm{v} = \left[ \begin{array}{cccc} c_\mathrm{v} & c_\mathrm{p} & k_\mathrm{v} & q \end{array} \right]^\top$ has unknown variable $q$.
If there exists an absolutely continuous function $\bm{g}$ which is invertible and satisfies $\bm{v} = \bm{g} \left( \bm{u} \right)$, then the joint PDF for $\bm{v}$ is computed as,
\begin{eqnarray}
f_{c_\mathrm{v},c_\mathrm{p},k_\mathrm{v},q} \left( c_\mathrm{v}, c_\mathrm{p}, k_\mathrm{v}, q \right) &=& f_{c_\mathrm{v},c_\mathrm{p},k_\mathrm{v},q_\mathrm{ini}} \left( \bm{g}^{-1} \left( \bm{v} \right) \right)|J|, \\
J &=& \mathrm{det} \left( \frac{\partial \bm{u}}{\partial \bm{v}} \right). \nonumber
\end{eqnarray}
In this study, the analytical solution eq.(\ref{eq:jointmodel}) that is obtained in Section~\ref{sec:solution} is the 4th element of the function $\bm{g}$.
Finally, by integrating $f_{c_\mathrm{v},c_\mathrm{p},k_\mathrm{v},q}$ over all parameters expecting $q$, the PDF of joint angle is obtained:
\begin{equation}
f_q \left( q \right) = \int_0^\infty \int_0^\infty \int_0^\infty f_{c_\mathrm{v},c_\mathrm{p},k_\mathrm{v},q} \left( c_\mathrm{v}, c_\mathrm{p}, k_\mathrm{v}, q \right) dc_\mathrm{v} dc_\mathrm{p} dk_\mathrm{v}.
\end{equation}

\begin{figure}
\centering
\subfloat[Joint 1.]{%
\resizebox*{4.5cm}{!}{\includegraphics{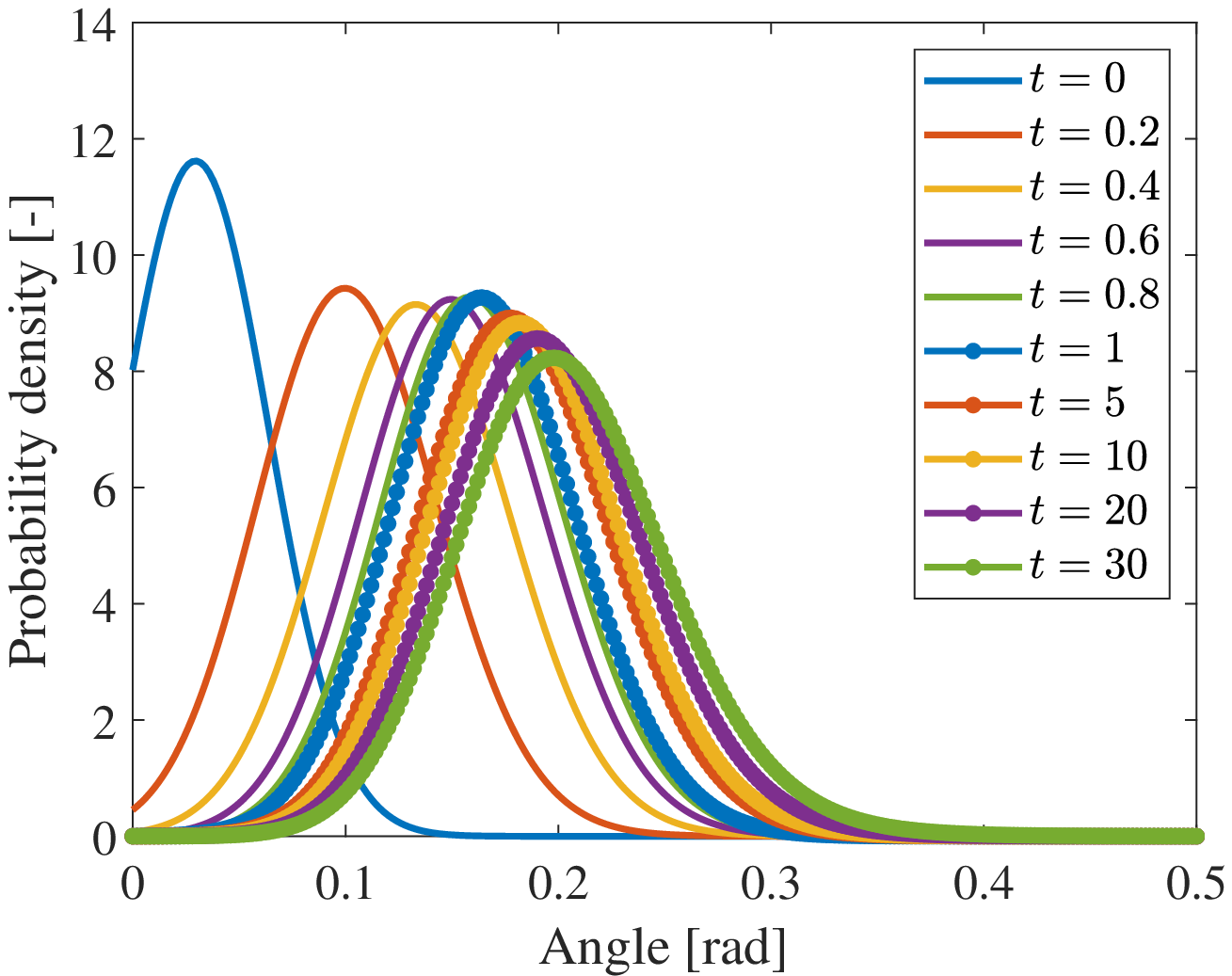}}}\hspace{5pt}
\subfloat[Joint 2.]{%
\resizebox*{4.5cm}{!}{\includegraphics{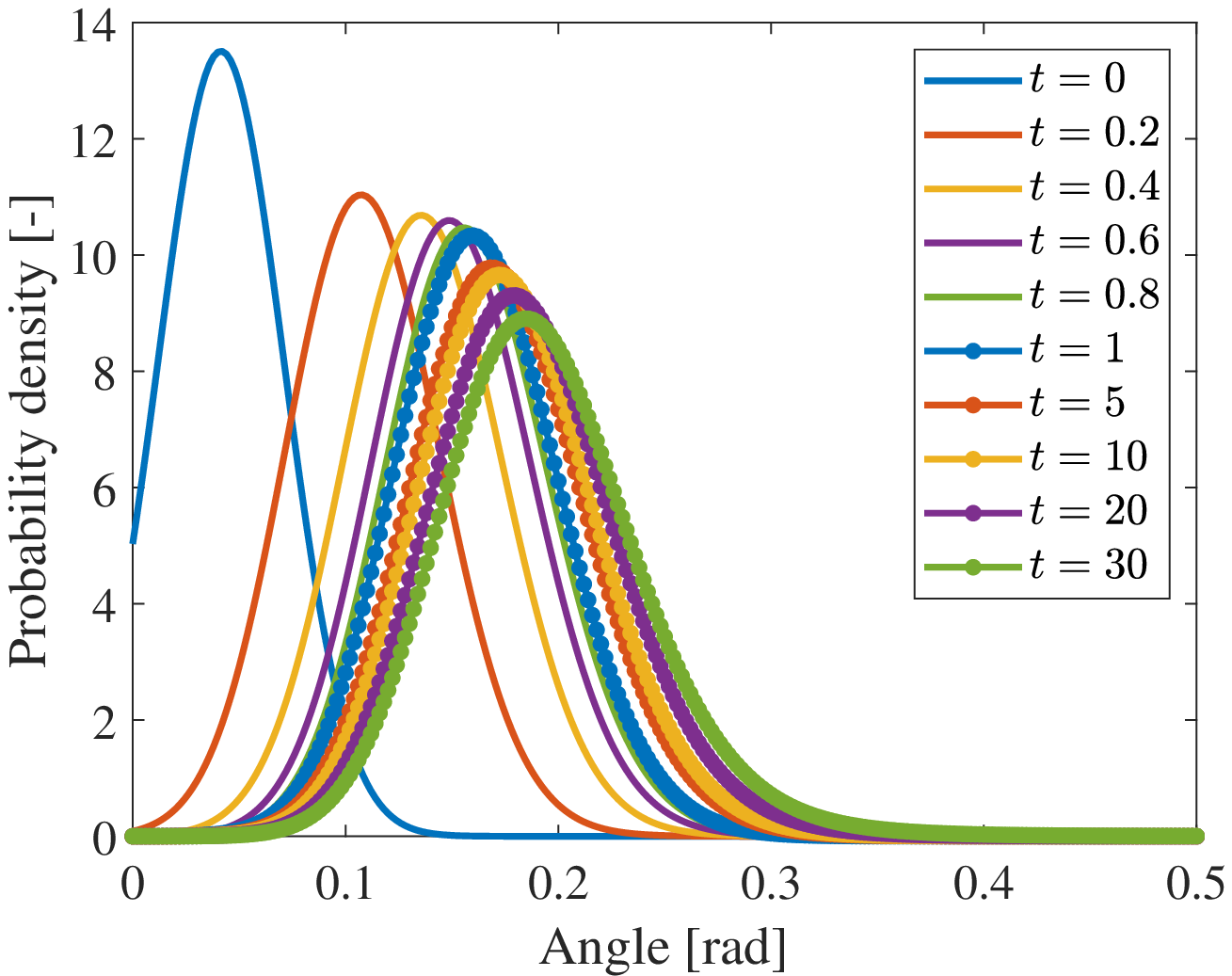}}}\hspace{5pt}
\subfloat[Joint 3.]{%
\resizebox*{4.5cm}{!}{\includegraphics{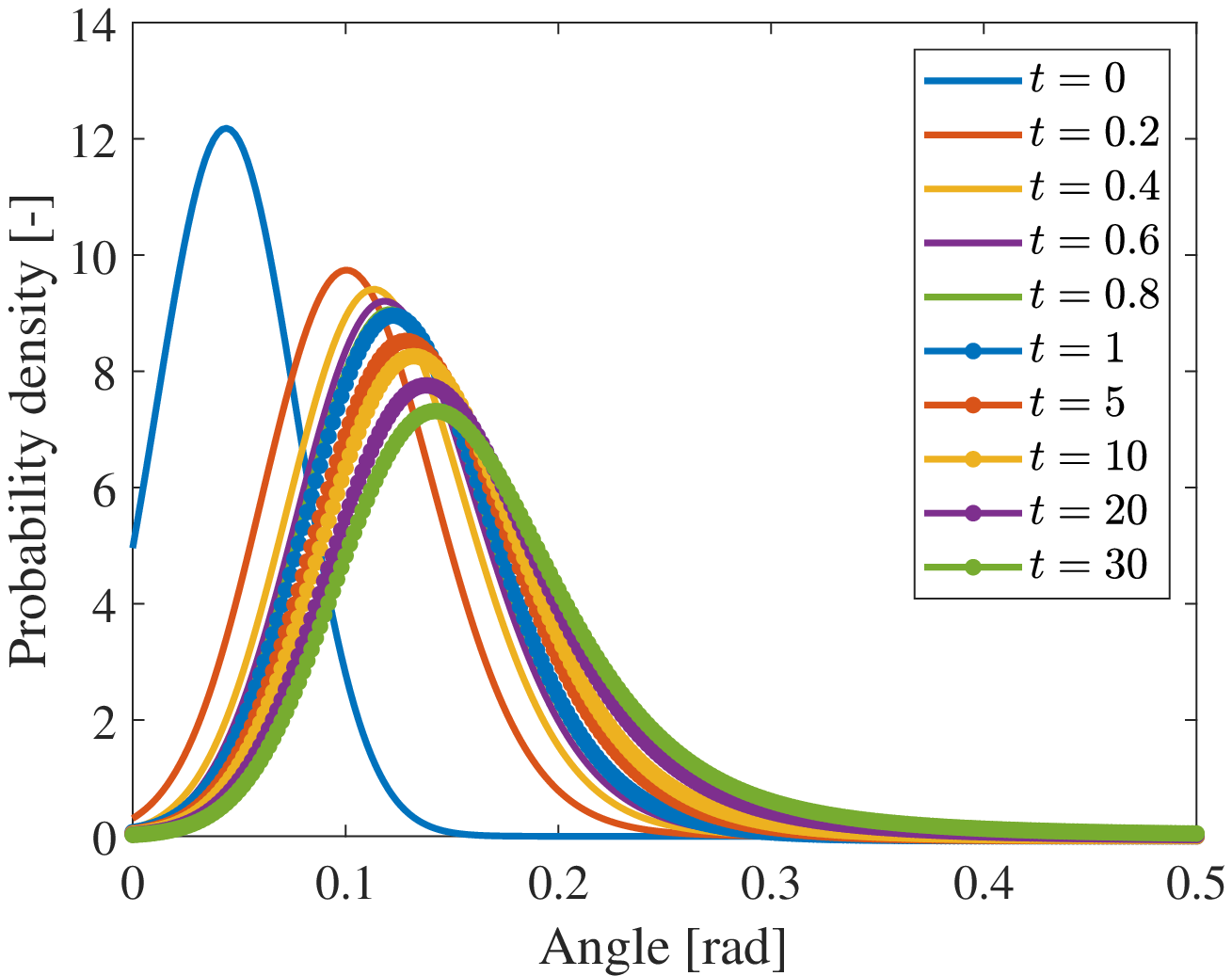}}}\hspace{5pt}
\caption{
The PDF of each joint displacement at each time step is shown.
} \label{fig:jointdisplacement}
\end{figure}

The PDF of each joint displacement is shown in Figure~\ref{fig:jointdisplacement}.
When $t = 0$, the shapes of the PDFs are equal to the distribution of the initial joint angle, then the peak value shifts to large with time evolution.
This means that the joint deforms due to its plasticity.
At the same time, the height of the peaks becomes smaller.
Since it is the PDF, this indicates that the variance becomes larger, in other words, the variability of the joint displacement is high with time development.

\begin{figure}
\centering
\subfloat[Joint 1 (macro view).]{%
\resizebox*{4.5cm}{!}{\includegraphics{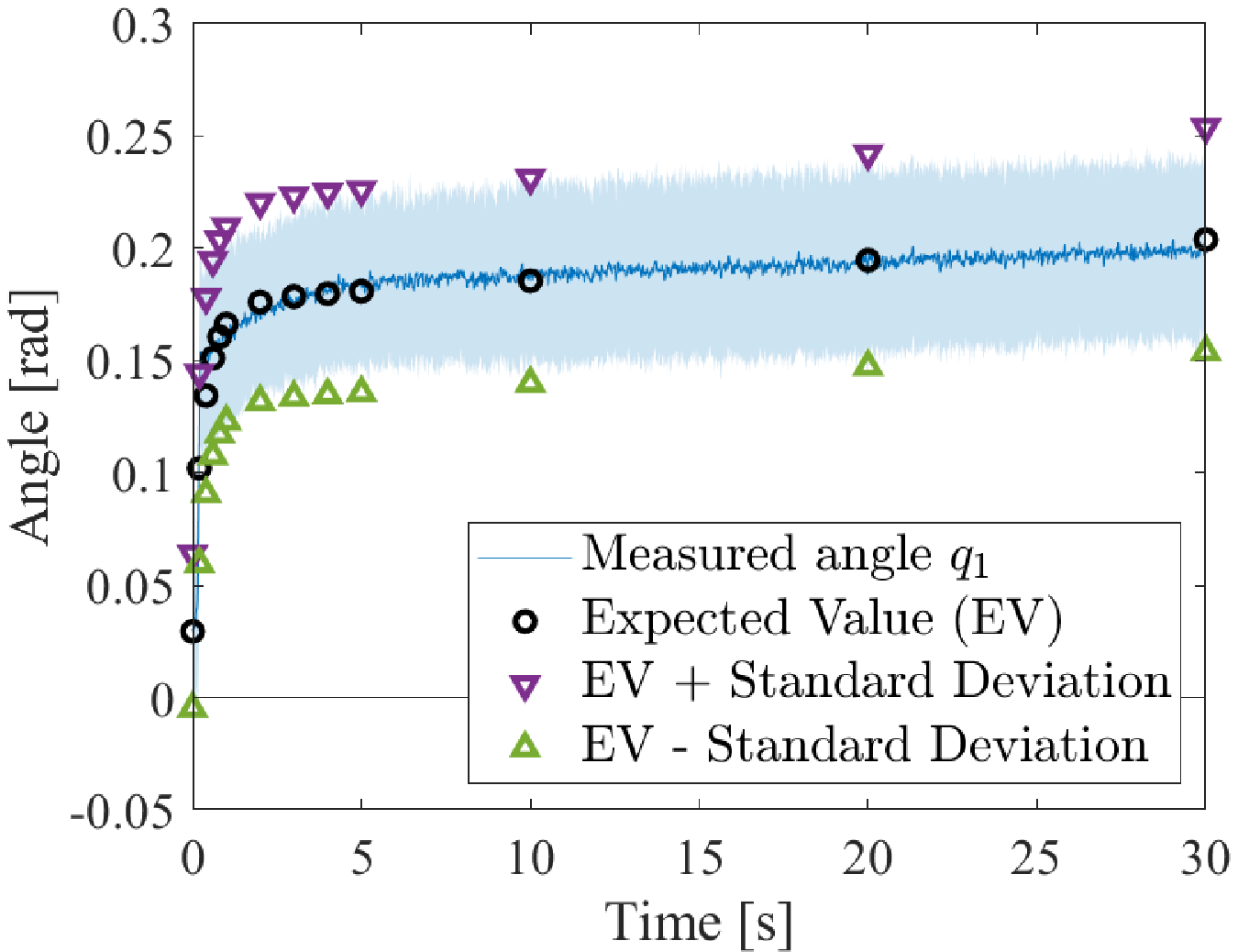}}}\hspace{5pt}
\subfloat[Joint 2 (macro view).]{%
\resizebox*{4.5cm}{!}{\includegraphics{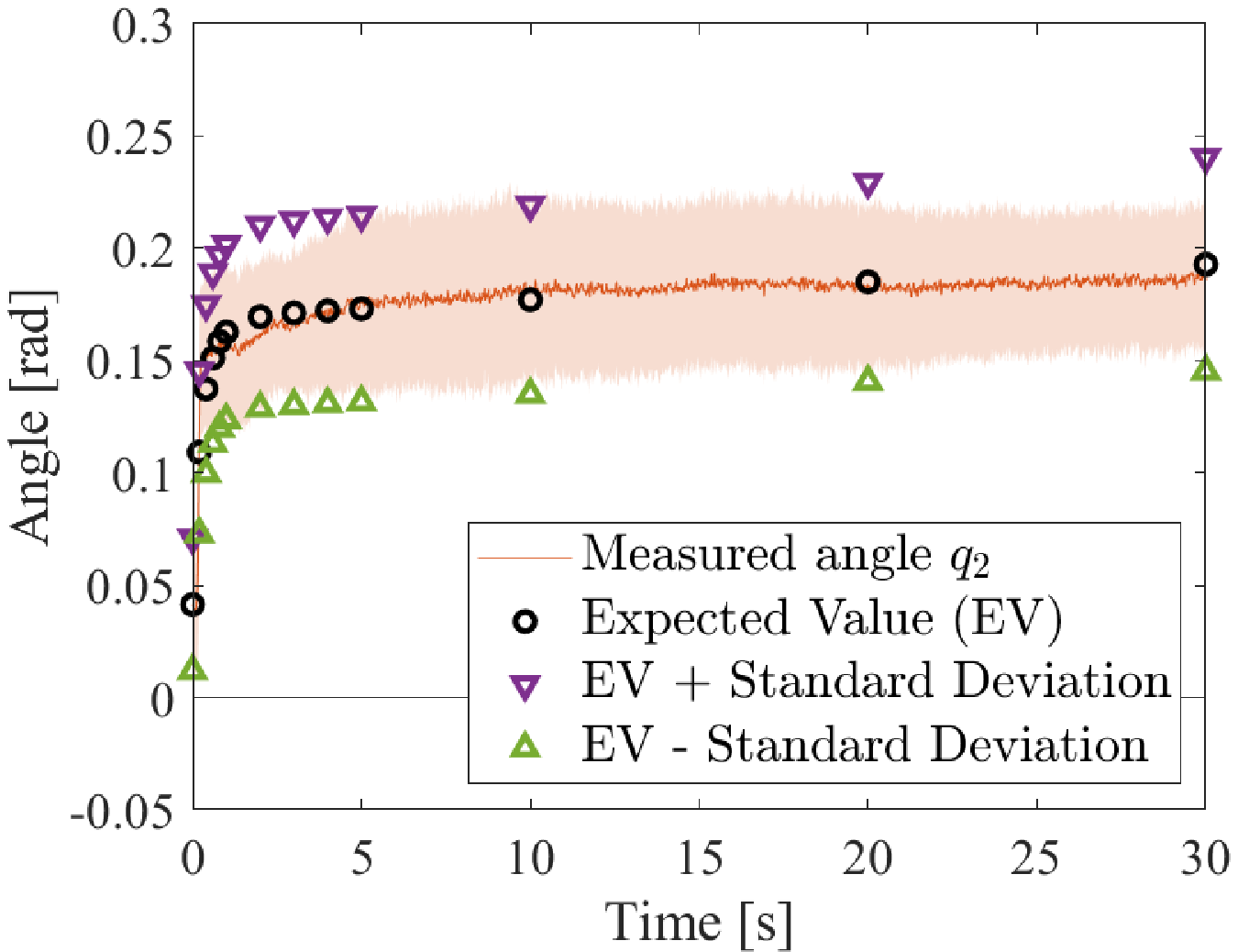}}}\hspace{5pt}
\subfloat[Joint 3 (macro view).]{%
\resizebox*{4.5cm}{!}{\includegraphics{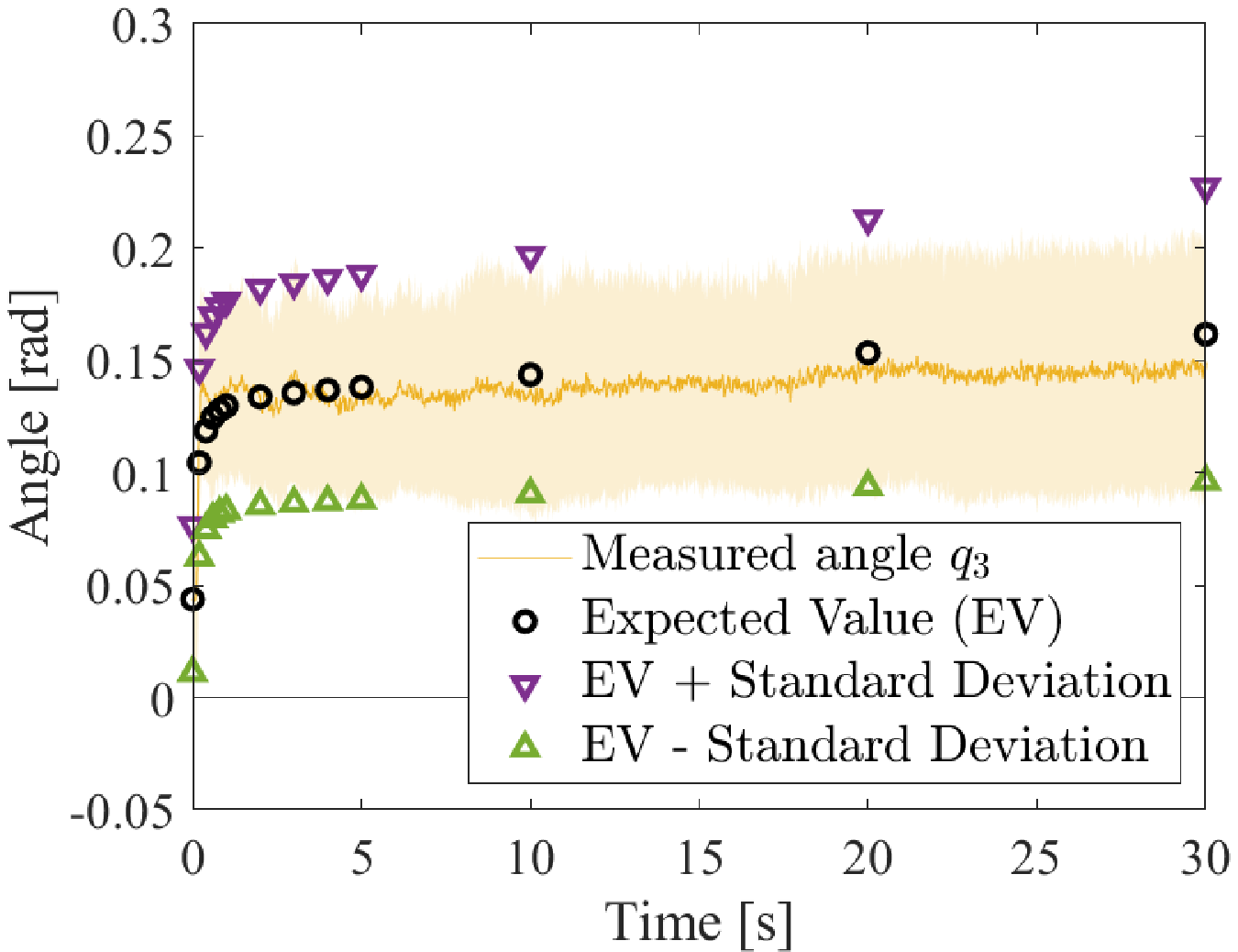}}}\hspace{5pt}
\subfloat[Joint 1 (micro view).]{%
\resizebox*{4.5cm}{!}{\includegraphics{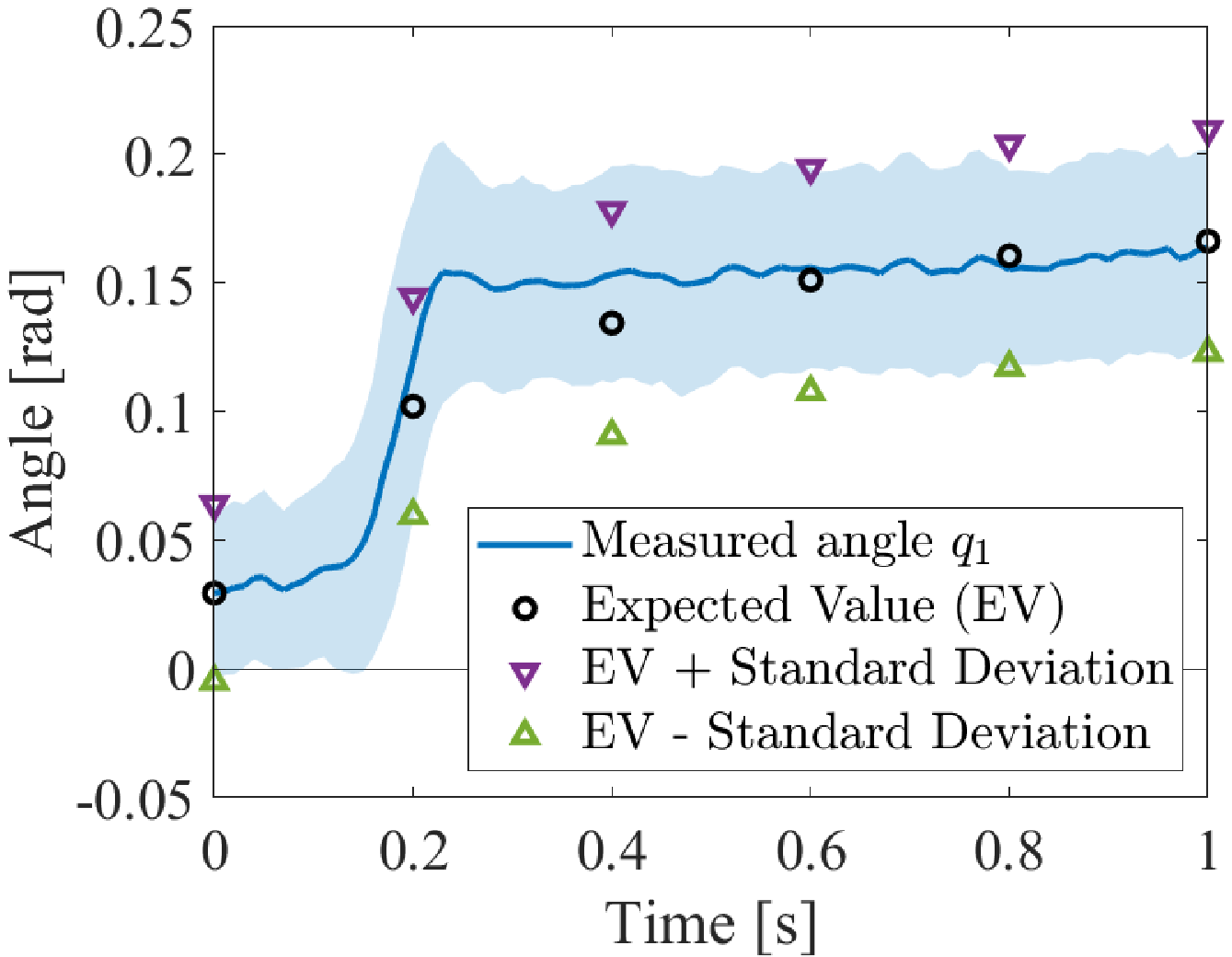}}}\hspace{5pt}
\subfloat[Joint 2 (micro view).]{%
\resizebox*{4.5cm}{!}{\includegraphics{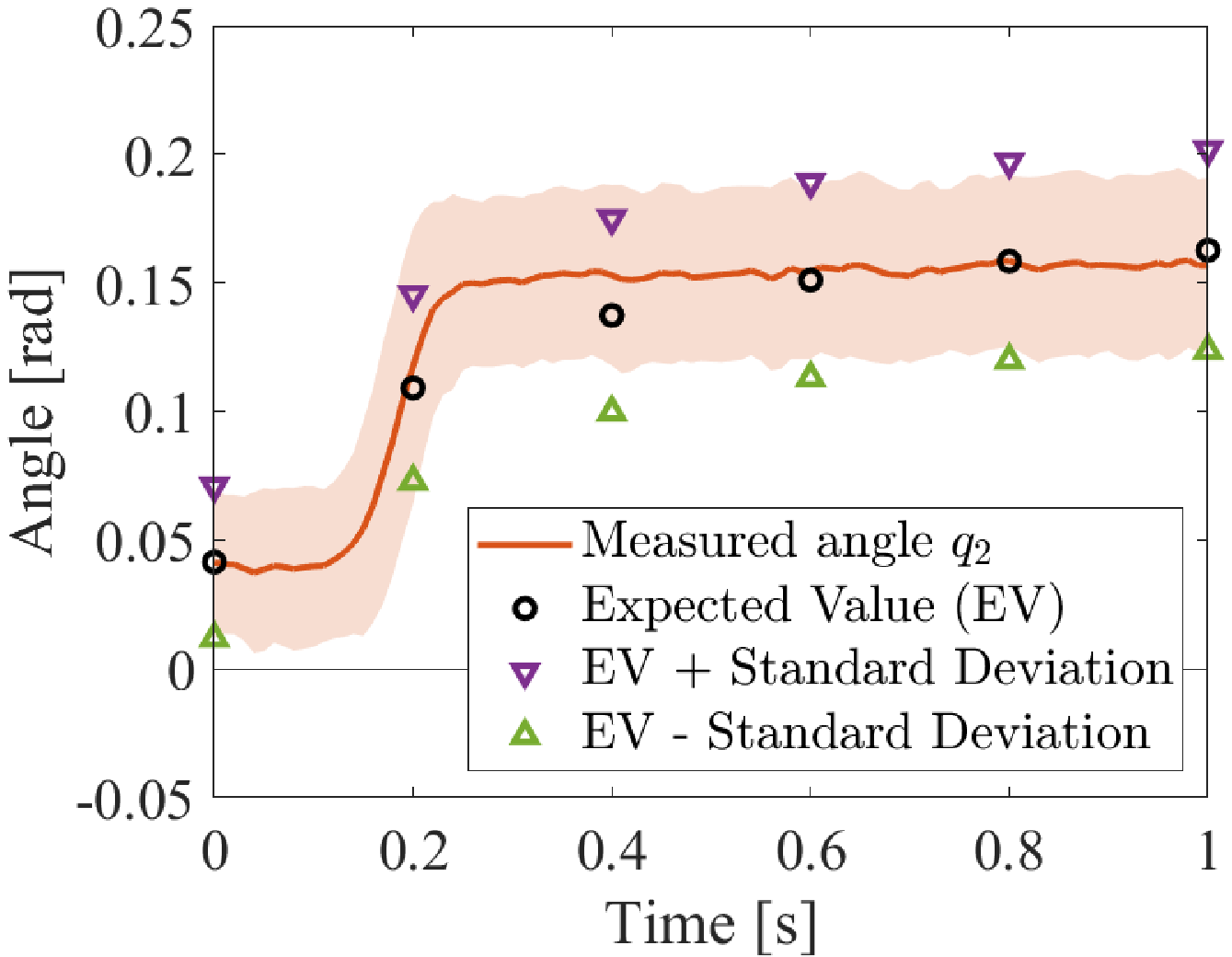}}}\hspace{5pt}
\subfloat[Joint 3 (micro view).]{%
\resizebox*{4.5cm}{!}{\includegraphics{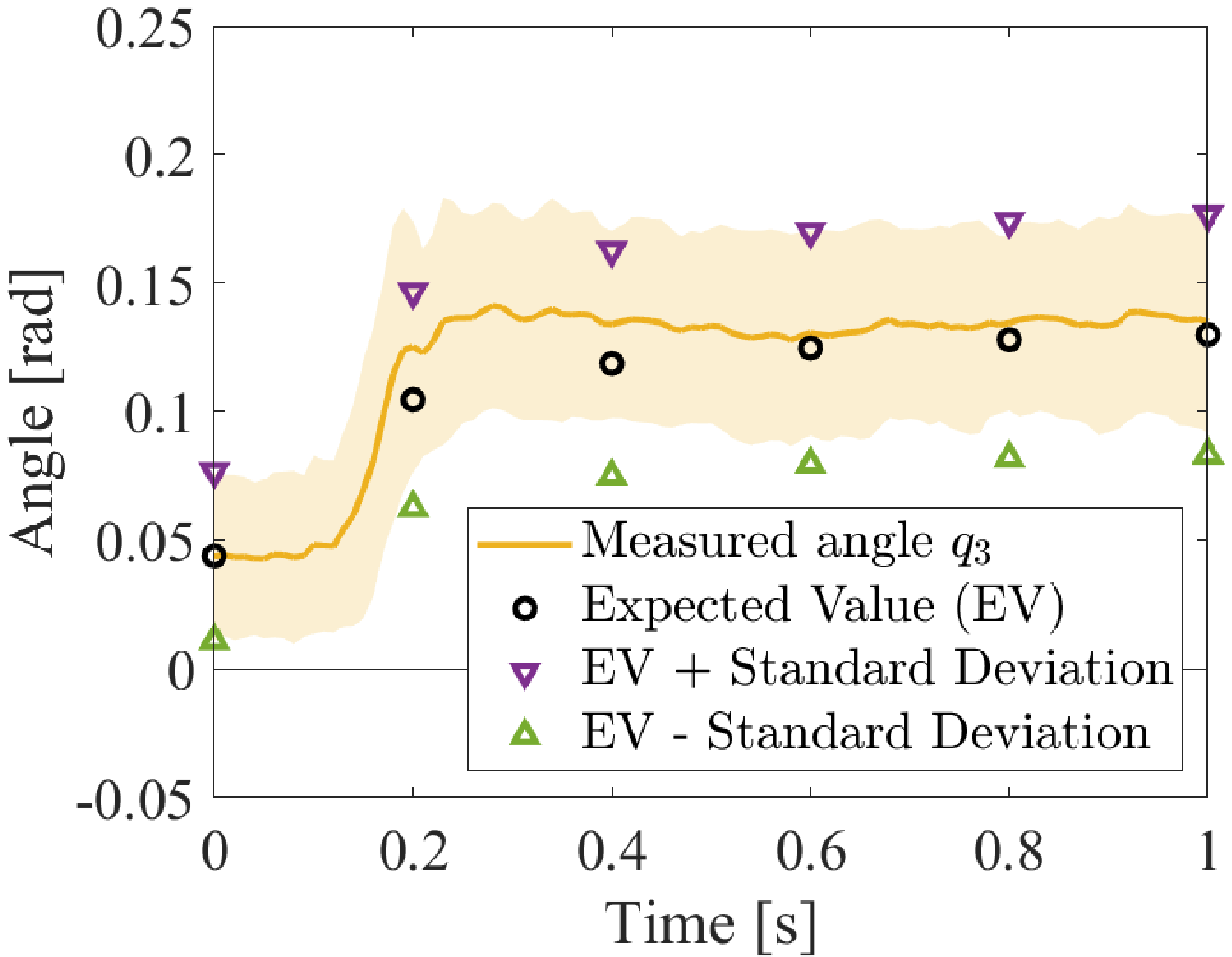}}}\hspace{5pt}
\caption{
The experimental results and analytical solutions are plotted.
The solid line and the colored band are the experimental average and its standard deviation respectively.
The dotted lines are stochastic values, expected value and standard deviation respectively.
(a) $\sim$ (c) show the whole deformaions, and (d) $\sim$ (f) show the trainsient responses.
} \label{fig:comparison}
\end{figure}

The comparison between the experimental results and the stochastic expression is shown in Figure~\ref{fig:comparison}.
The solid line is the average joint displacement that is measured in the experiment, and the colored band is the average $\pm$ standard deviation (SD).
Also, the dotted line means the expected value (EV) and the EV $\pm$ standard deviation that is obtained from stochastic analysis.
Both results are similar, which means that we can estimate the joint deformation from the distribution of each parameter.
The SDs of experimental and analytical results shows similar transition at each joint.
This means that the variability of the joint motion can be estimated using the stochastic analysis.
In addition, the creep behavior is shown in Figure~\ref{fig:comparison}, analytical EV and SD correspond to the experimental result.
Our model shows the creep behavior at each joint, which is mainly confirmed in Figure~\ref{fig:comparison}(a) $\sim$ (c).
The transient response also appears in Figure~\ref{fig:comparison}(d) $\sim$ (f), which implies that the proposed 3-elements viscoelastic joint model improves the shortcoming of the Maxwell model.
Though our approach shows adequate results at each joint, there is a little gap in joint 3, shown in Figure~\ref{fig:comparison}(c).
We investigate the cause using sensitivity analysis in the next chapter.

\subsection{Sensitivity analysis}

To investigate which parameter affects the variability of the joint displacement in Figure~\ref{fig:jointdisplacement}, the sensitivity analysis is performed.
The Sobol method is one of the most representative methods of global sensitivity analysis based on the variance.
In this method, Sobol indices are calculated for each parameter or the combination of parameters, and they indicate how much the parameter influences the total variance.
Here is a brief description, for the seek of details, please see \cite{Sobol1990}.
In general, the model function $f \left( \bm{x} \right)$ is defined in the supercubic space $\Omega^k = \left\{ \bm{x}|0\leq x_i \leq 1;i=1, \cdots, k\right\}$.
Then this function is decoupled as shown below:
\begin{equation}
f \left( \bm{x} \right) = f_0 + \sum_{i=1}^k f_i \left( x_i \right) + \sum_{1 \leq i < j \leq k} f_{i,j} \left( x_i, x_j \right) + \cdots + f_{1,2,...,k} \left( x_1, x_2, ..., x_k \right).
\end{equation}
This expansion is uniquely determined if $f \left( \bm{x} \right)$ is integrable over $\Omega^k$, and each term is obtained as,
\begin{eqnarray}
f_0 &=& \int_{\Omega^k} f \left( \bm{x} \right) d\bm{x}, \\
f_i \left( x_i \right)&=& \int_0^1 \cdots \int_0^1 f \left( \bm{x} \right) d\bm{x}_{\sim i} - f_0, \label{eq:decouple} \\
f_{i,j} \left( x_i, x_j \right)&=& \int_0^1 \cdots \int_0^1 f \left( \bm{x} \right) d\bm{x}_{\sim i,j} - f_i \left( x_i \right) - f_j \left( x_j \right) - f_0.
\end{eqnarray}
Here, $d\bm{x}_{\sim i}$ means to integrate all elements of $\bm{x}$ except $x_i$.
A higher term is obtained by the same procedure.
Then the sensitivity indices $S_{i_1,\cdots,i_s}$ are defined for the total variance $D$ and the partial variances $D_{i_1,\cdots,i_s}$ as follows:
\begin{eqnarray}
D &=& \int_{\Omega^k} f^2 \left( \bm{x} \right) d\bm{x} - f_0^2, \label{eq:variance} \\
D_{i_1,\cdots,i_s} &=& \int_0^1 \cdots \int_0^1 f_{i_1,\cdots,i_s}^2 \left( x_{i_1},\cdots,x_{i_s} \right) dx_{i_1}\cdots dx_{i_s}, \label{eq:partialvariance} \\
S_{i_1,\cdots,i_s} &=& \frac{D_{i_1,\cdots,i_s}}{D} \hspace{10pt}  \left( 1 \leq i_1 < \cdots < i_s \leq k \right).
\end{eqnarray}
$S_i$ indicates how $x_i$ affects the variance of $f \left( \bm{x} \right)$, called the main effect.
$S_{i,j}$ expresses the interaction effect of the variables $x_i$ and $x_j$, and higher terms are equal.
The sum of all sensitivity indices is 1.

In this study, the model function is eq.\eqref{eq:jointmodel}, and the parameters $c_\mathrm{v}, c_\mathrm{p}, k_\mathrm{v}, q_\mathrm{ini}$ aren't on the supercubic space.
Since each parameter is expressed as stochastic distribution, we can derive the relationship between the uniform variable and the distributed variable using the inversion method.
If the cumulative distribution function of parameter $x$ is defined as $F_x \left( x \right)$, then $u_x = F_x \left( x \right), 0 \leq u_x \leq 1$ is satisfied.
Finally, $y = f \left( \bm{x} = \bm{F}^{-1} \left( \bm{u} \right) \right) = g \left( \bm{u} \right)$ becomes the new model function.
Also, we only discuss the first-order Sobol indices $S_i$ because their sum equals 1, which means that the main effect of each parameter is large enough for the variability of each joint angle.
Sobol indices are computed using the Monte-Carlo integral instead of solving eq.(\ref{eq:variance}) and (\ref{eq:partialvariance}) directly, since eq.(\ref{eq:decouple}) can not be obtained analytically.
The details are found in \cite{Homma1996}.

\begin{figure}
\centering
\subfloat[Joint 1.]{%
\resizebox*{4.5cm}{!}{\includegraphics{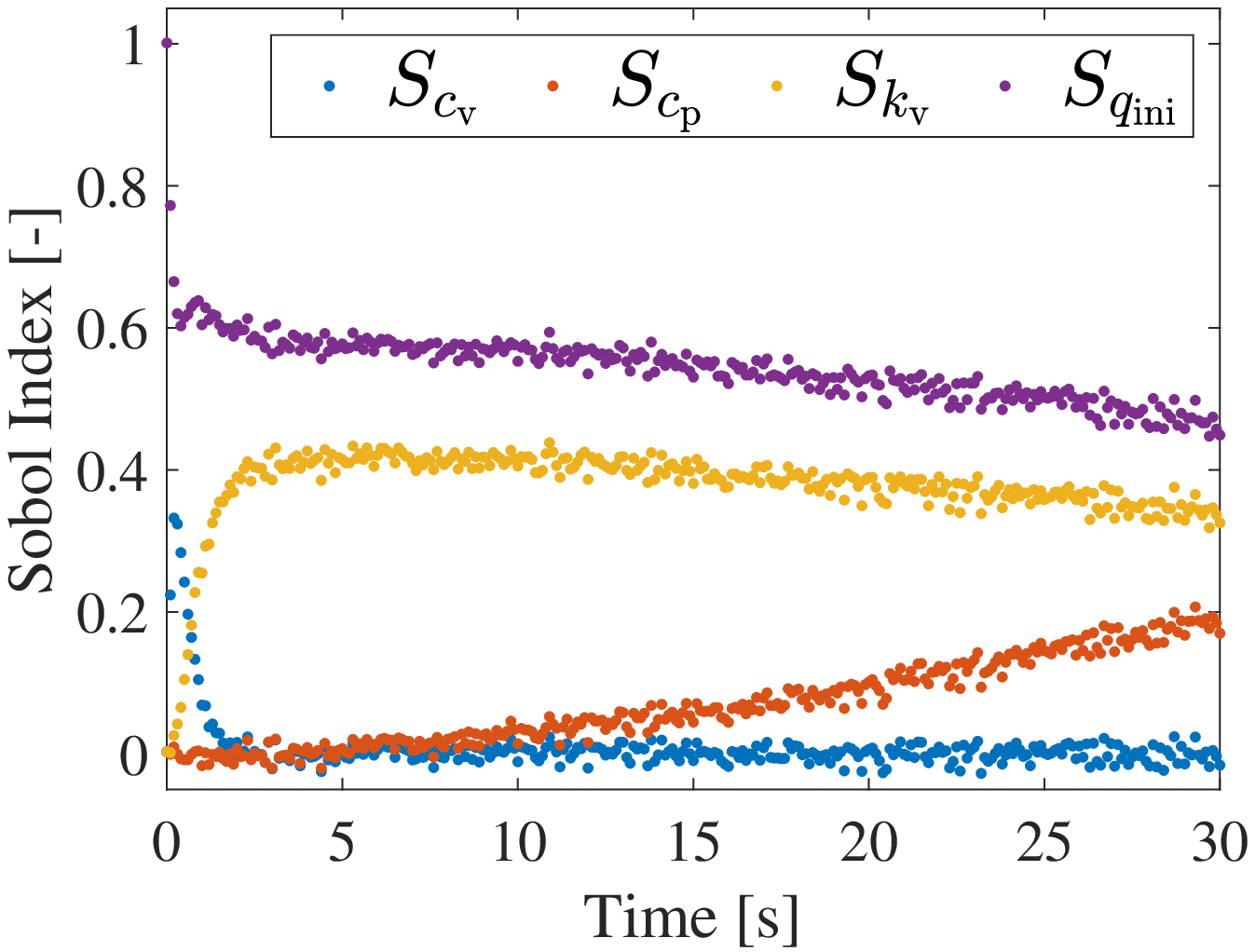}}}\hspace{5pt}
\subfloat[Joint 2.]{%
\resizebox*{4.5cm}{!}{\includegraphics{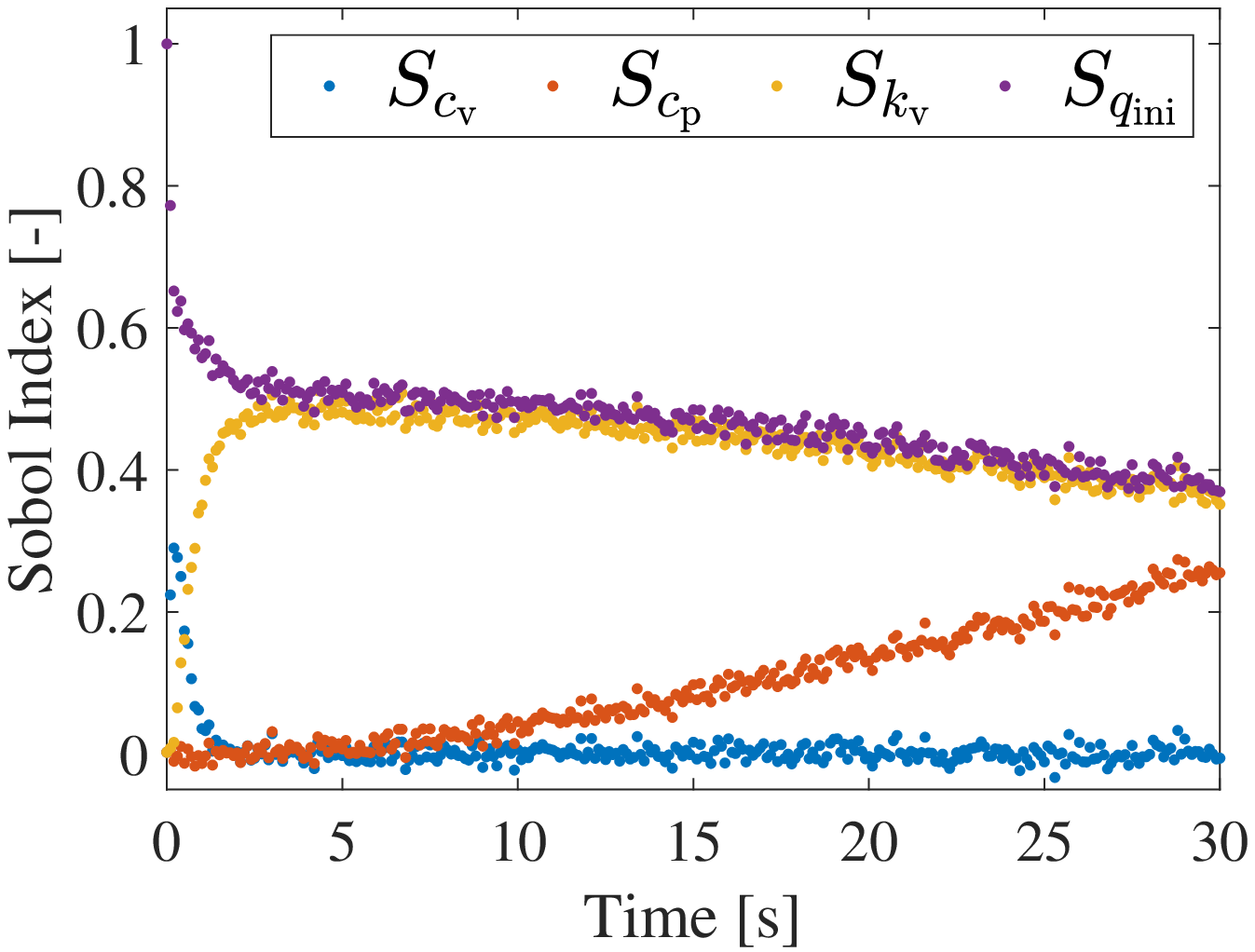}}}\hspace{5pt}
\subfloat[Joint 3.]{%
\resizebox*{4.5cm}{!}{\includegraphics{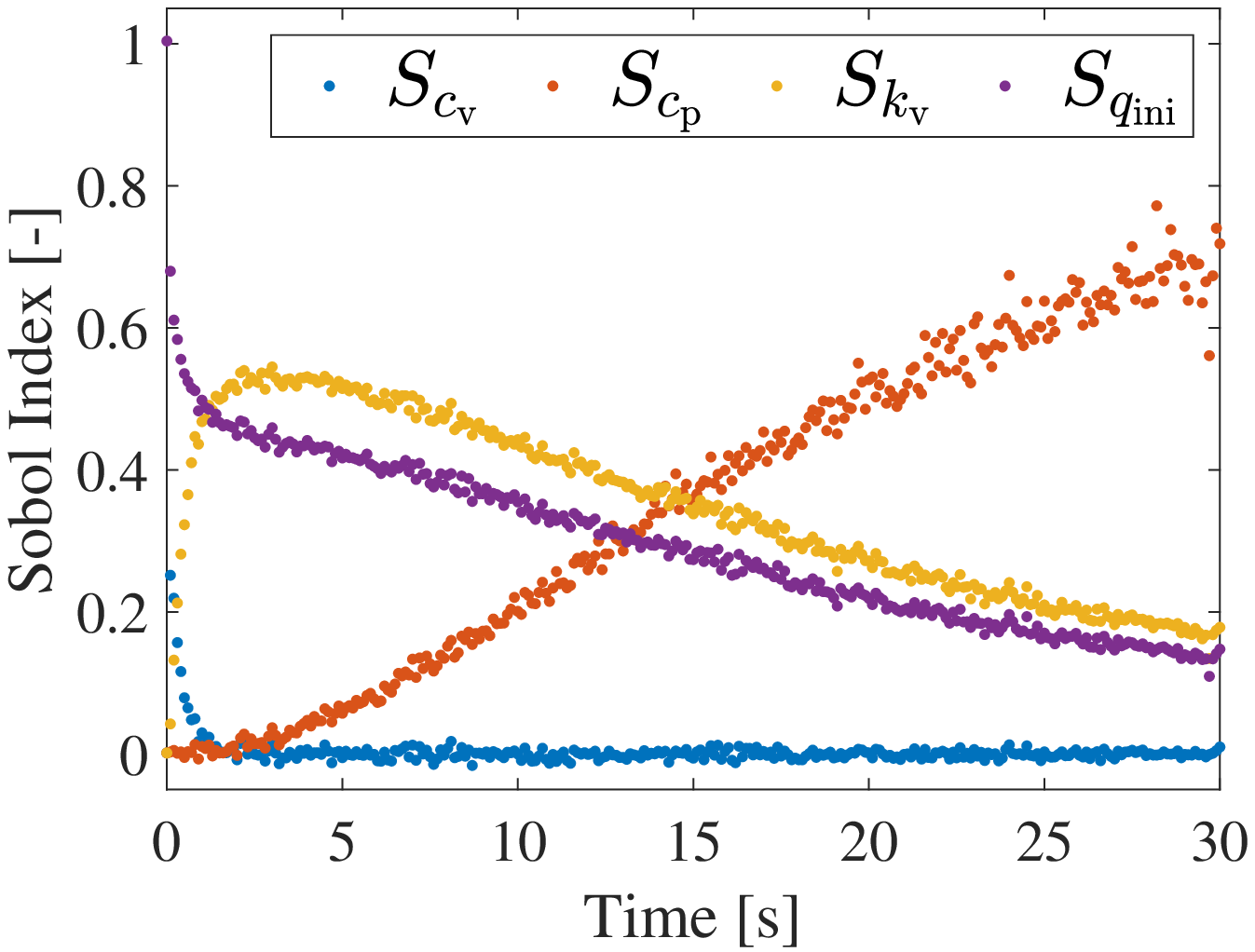}}}\hspace{5pt}
\caption{Sobol indices with time development.} \label{fig:sobol}
\end{figure}

Figure~\ref{fig:sobol} shows the resulting Sobol indices over time.
Each index is computed in 0.1-second time steps.
It shows a qualitatively reasonable result.
When $t = 0$, there is only the effect of the initial angle, it decreases with time.
The effect of $c_\mathrm{v}$ appears early when an activation torque occurs.
$S_{k_\mathrm{v}}$ also increases at first, then decreases slowly.
$S_{q_\mathrm{p}}$ increases after $S_{c_\mathrm{v}}$ disappears, which means the deformation phase shifts from elastic to plastic.

Considering the difference between joints, the effect of series viscosity $c_\mathrm{p}$ is greater at joint 3 than at other joints with time transition.
Because these indices mean relative effects, we consider this phenomenon from all distributions of parameters shown in Figure~\ref{fig:dist}.
The distribution of parallel elasticity $k_\mathrm{v}$ at joint 3 is greatly different from the other 2 joints, and $k_\mathrm{v}$ of joint 3 can take a wide range of values stochastically.
This may cause the decrease of the effect of $k_\mathrm{v}$ and lead to the growth of $c_\mathrm{p}$ as a result.
Furthermore, this phenomenon can be explained from the model function eq.(\ref{eq:jointmodel}).
The parallel viscosity $c_\mathrm{v}$ appears only in the exponential, so its effect decreases exponentially.
With time transition, the terms of $c_\mathrm{p}$, $k_\mathrm{v}$ and $q_\mathrm{ini}$ exist.
Since the term of $c_\mathrm{p}$ is first-order of time $t$, its relative effect increases, and that of $k_\mathrm{v}$ and $q_\mathrm{ini}$ disappear as time goes by.

\section{Discussion}\label{sec:discussion}

We obtain the analytical solution in Section~\ref{sec:solution} under some assumptions.
Figure~\ref{fig:comparison} shows that the stochastic analysis offers almost the same result compared with the experiment.
This means that our assumptions are enough for this experiment.

By these stochastic analyses, we can treat uncertainties that soft robots contain in their properties.
We show how much the angle displacement varies with time transition qualitatively in Figure~\ref{fig:jointdisplacement}.
It is revealed that the distribution of joint 3 changes more drastically than the other 2 joints, which comes from the trend that the distributions of viscoelastic parameters are low and wide at joint 3.
Quantitatively, we show the sensitivity indices in Figure~\ref{fig:sobol}.
They are useful when we design soft robots.
In this case, for example, we should pay attention to the viscoelasticity $c_\mathrm{v}$ and $k_\mathrm{v}$ when we use the soft finger for a short time and to the series viscosity $c_\mathrm{p}$ for a long time.
Especially we can evaluate the parameters' effects with exact the running time.

Though the proposed model is established from the soft finger whose shape is suitable for the lumped parameterized model, it can be applied to many kinds of soft robots.
For example, our model can be used for continuum manipulators by increasing links and joints.
In addition, our model is extendable by changing the joint viscoelastic component.
We introduce a 3-elements model to express the creep behavior, and other types of viscoelastic combinations such as the general Maxwell, the general Voigt model, and so on can be meaningful for more detailed analysis.
For such cases, the stochastic analyses used in this study are possible if an analytical solution can be derived.

However, these analyses are based on the analytical solution (\ref{eq:jointmodel}).
As we mention, it is hard to derive it for general input, so the proposed method cannot be utilized for every case.
The appropriate assumptions or constraints are required.

\section{Conclusion}

In this study, the stochastic approach is proposed for modeling a soft finger.
The dynamics is described as the lumped parameterized model with joint viscoelasticity.
The linear combination of the parallel viscoelastic element and the series damper is derived as the joint model, which makes it possible to express the creep phenomenon of a soft finger.
Through the parameter estimation experiment, the viscoelastic parameters are obtained as distributed ones to explain the poorly-reproducible behavior.
Furthermore, the evolution of joint angles is calculated by the random variable transformation, and the stochastic analysis and the experimental result show similar characteristics.
In addition, the global sensitivity analysis is performed using the Sobol method, which shows a reasonable result in terms of the model mechanism.

As a plan, the control scheme based on the stochastic model will be considered.
First of all, we try to use our model as an observer in control.
We aim to realize dexterous and safe manipulations by making use of the characteristics of the soft finger.

\section*{Acknowledgements}

We greatly appriciate all fundings above.
Also, we would like to thank the members of HCR lab. for useful discussions.

%
%
%
\section*{Funding}

This work was supported by the Grant-in-Aid for Scientific Research (A) No. 20H00610 of Japan Society for the Promotion of Science (JSPS).

\end{document}